\documentclass[11pt,a4paper,nonumbering]{ds}

\usepackage[authoryear,sort&compress,round]{natbib}
\usepackage{graphicx}
\usepackage{booktabs}
\usepackage{multirow}
\usepackage{array}
\usepackage{tabularx}
\usepackage{makecell}
\usepackage{float}
\usepackage{dblfloatfix}
\usepackage{caption}
\usepackage{xcolor}
\usepackage{enumitem}
\usepackage{listings}
\usepackage{url}

\renewcommand{\today}{}
\reportnumber{001}

\definecolor{tablegray}{gray}{0.94}
\definecolor{codegray}{gray}{0.97}

\newcommand{\same}{Same-set Det. Val.}
\newcommand{\cross}{Cross-set Det. Test}

\newcommand{\best}[1]{\textbf{#1}}
\newcommand{\second}[1]{\underline{#1}}
\newcolumntype{C}{>{\centering\arraybackslash}X}
\newcolumntype{Y}{>{\raggedright\arraybackslash}X}

\title{\centering LocAnyMed: Vision-Language Grounding for Multimodal Medical Images}

\makeatletter
\def\@author{
\begin{center}
\normalsize
Zihan Wang$^{1,\dagger}$,
Tong Liu$^{1,\dagger}$,
Zhiwei Wang$^1$,
Tao Huang$^1$,
Wentao Jiang$^1$,
\\[2pt]
Sihan Ma$^2$,
Shanshan Ye$^3$,
Xiaohui Yang$^4$,
Jing Zhang$^{1,*}$
\\[5pt]
{\footnotesize
\renewcommand{\arraystretch}{1.05}
\begin{tabular}{c}
$^1$School of Computer Science, Wuhan University, Wuhan, China\\
$^2$College of Computing and Data Science,
Nanyang Technological University, Singapore\\
$^3$Australian Artificial Intelligence Institute,
School of Computer Science,\\
University of Technology Sydney, Sydney, Australia\\
$^4$Henan University, Kaifeng, China
\end{tabular}}
\\[5pt]
{\small
$^*$Corresponding author:
\texttt{jingzhang.cv@gmail.com}}
\end{center}
}
\makeatother

\correspondingauthor={}

\fancypagestyle{firststyle}{
    \fancyhead[R]{}
    \fancyhead[L]{{\footerfont\itshape\monthyeardate\today}}
    \fancyhead[C]{}

    \fancyfoot[L]{
        \footerfont
        $^\dagger$ These authors contributed equally to this work.
        $^*$ Corresponding author.
    }
    \fancyfoot[R]{}
    \fancyfoot[C]{\footerfont \thepage}
}


\begin{abstract}
Medical visual grounding connects free-form clinical queries to spatial evidence in medical images and is an important component of interpretable medical artificial intelligence. However, general-purpose grounding models are predominantly trained on natural images, while existing medical localization resources remain fragmented across imaging modalities, datasets, and task formulations.  To address this gap, we construct LocAnyMed-200K, a multimodal medical visual grounding dataset containing approximately 200K image-query-answer examples across computed tomography, optical medical imaging, ultrasound, and X-ray. We harmonize heterogeneous detection and localization resources into a unified free-form instruction format that supports one or multiple bounding boxes, point coordinates, and no-target outputs for negative queries. Full-parameter fine-tuning of LocateAnything-3B on LocAnyMed-200K improves F1@IoU 0.50 from 10.64 to 85.59 on a held-out evaluation split, demonstrating that large-scale domain-specific supervision can equip a general grounding model with effective medical localization capabilities. Beyond spatial coordinates, a clinically interpretable grounding system should also communicate the evidence supporting its prediction. We therefore derive LocAnyMed-CoT-20K, a rationale-augmented subset that connects anatomical context, visual observations, and spatial conclusions through structured reasoning and further improves cross-source generalization through fine-tuning. Together, these resources provide a unified foundation for studying both localization accuracy and rationale quality across heterogeneous medical imaging modalities. The code is publicly available at https://github.com/MiliLab/LocAnyMed.

\end{abstract}

\begin{document}
\maketitle

\section{Introduction}

\begin{figure*}[t]
    \centering
    \includegraphics[
        width=\textwidth,
        trim={4.5cm 2cm 4cm 1.4cm},
        clip
    ]{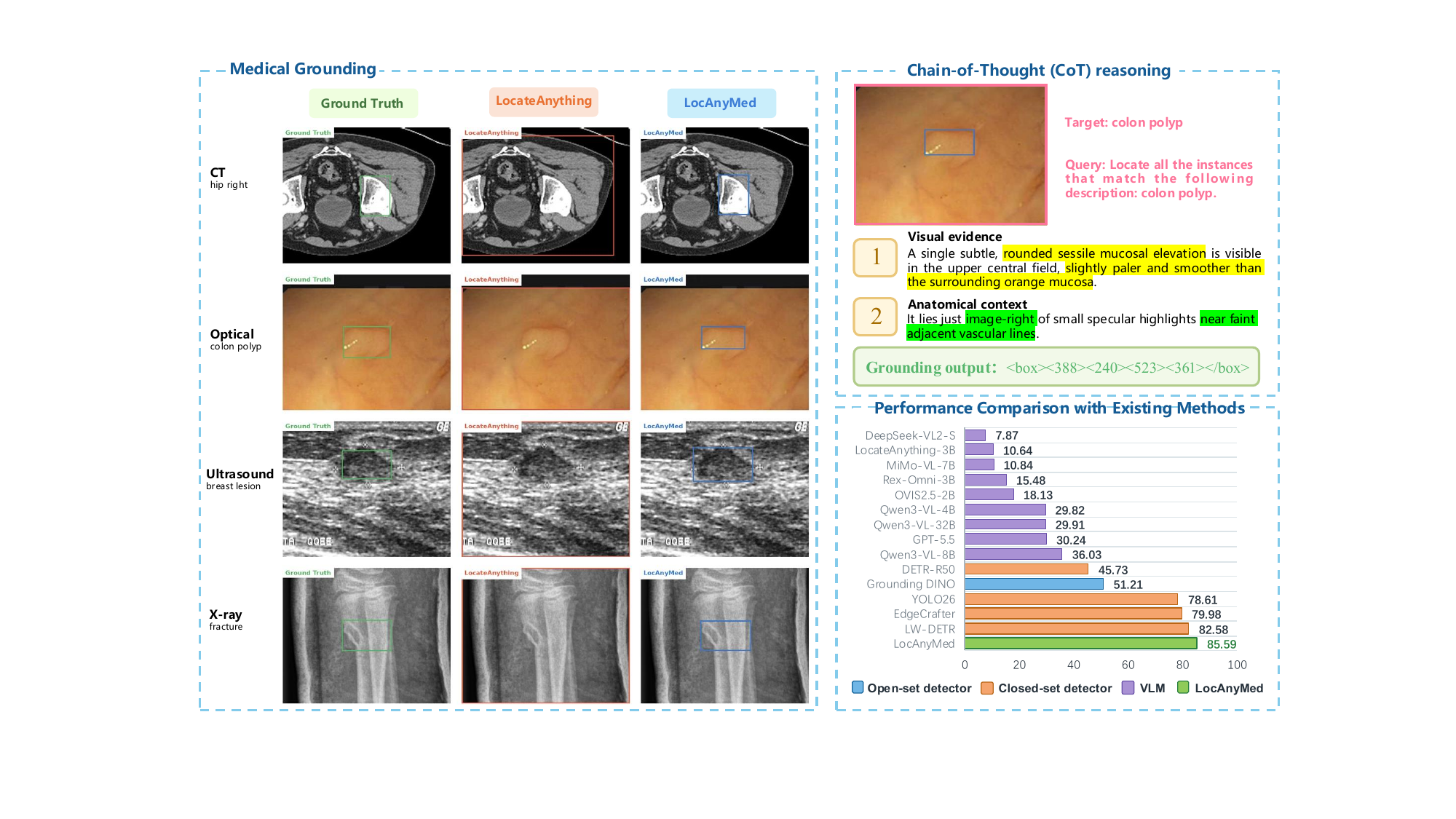}
    \caption{
    Overview and representative results of LocAnyMed. \textbf{Left:} Qualitative grounding results across four medical imaging modalities. \textbf{Top right:} A visual-evidence rationale example. \textbf{Bottom right:} F1@IoU 0.50 comparison on Same-set Det. Val., where LocAnyMed achieves the best performance.}
    \label{Figure 1}
\end{figure*}

Grounding is a fundamental task in medical image analysis. It identifies regions associated with anatomical structures, lesions, and abnormal findings, and supports a wide range of diagnostic and quantitative tasks~\citep{nndetection2021}. Most medical detectors are developed for predefined categories and task-specific datasets. Although effective in closed-set settings, they cannot readily localize targets specified by open-ended clinical queries~\citep{nndetection2021}. Medical visual grounding provides a more flexible formulation by mapping a free-form text query to its corresponding image region~\citep{medrpg2023,umedground2025}. However, unified open-ended grounding across CT, optical images, ultrasound, and X-ray remains underexplored.

Recent vision-language models have extended open-ended grounding to general-domain images. LocateAnything provides a strong example by unifying detection and grounding through Parallel Box Decoding~\citep{locateanything2026}. However, it is not designed or trained for medical imaging. Transferring such a model is challenging because medical targets often exhibit low contrast, weak boundaries~\citep{medsam2024}, and large variations in scale and appearance. The problem becomes more difficult across imaging modalities. CT, optical imaging, ultrasound, and X-ray are generated by different physical processes and exhibit distinct visual characteristics~\citep{biomedparse2025}. Consequently, a general-domain grounding model cannot reliably localize medical targets without domain-specific data and adaptation.

To equip LocateAnything with medical grounding capability, we construct a unified dataset of approximately 200,000 samples spanning CT, optical imaging, ultrasound, and X-ray. Each sample is converted into an open-ended grounding format with bounding boxes, points, or empty outputs for negative queries. We then perform full-parameter supervised fine-tuning of LocateAnything-3B on this dataset~\citep{locateanything2026}. As shown in Figure~\ref{Figure 1}, on the held-out evaluation set, its detection F1@IoU 0.50 increases from 10.64 to 85.59, demonstrating the effectiveness of large-scale medical-domain adaptation. However, this domain-level adaptation does not explicitly model the substantial visual differences among the four imaging modalities.

MoonViT applies the same visual FFN to all imaging modalities, which may limit adaptation to their distinct visual characteristics~\citep{m4oe2024,medmoe2025}. We therefore decompose the output projection of each visual FFN into a shared branch and four modality-specific experts. The known modality activates one expert together with the shared branch, preserving the original active output width. We evaluate
several shared-to-expert ratios to balance shared representation and modality specialization.

Accurate localization alone does not reveal how a model connects a query to its image observations. A coordinate-only response states where a target is, but not how the model arrives at that location. This limitation motivates chain-of-thought (CoT) supervision for visual grounding, in which intermediate textual reasoning is generated before the spatial prediction~\citep{rexthinker2026,medclm2025}. We therefore distill approximately 20,000 samples from the full dataset and reorganize their labels into a CoT-augmented grounding format. The model is fine-tuned to produce a concise localization rationale followed by the target coordinates.

Together, these components form LocAnyMed, a unified framework for open-ended grounding across CT, ultrasound, X-ray, and optical images. Our contributions are twofold.

\begin{itemize}
    \item First, we construct a unified medical grounding corpus containing 209,910 image–query–answer records, covering box localization, point localization, and absent-target grounding. Using this corpus, we fully fine-tune LocateAnything-3B and introduce modality-aware shared-specific visual experts. 
    \item Second, we curate a rationale-augmented subset of 20,000 detection records through visual-evidence rationale distillation. This subset supervises the model to explain the visible evidence before producing a location or an explicit no-target response. 
\end{itemize}

\section{Related Work}

\textbf{Vision-Language Grounding in Medical Images.} Medical object localization has traditionally relied on closed-set detectors such as DeepLesion~\citep{deeplesion2018} and nnDetection~\citep{nndetection2021}, which predict predefined lesions or anatomical structures. General-domain grounding models instead connect detection with language. MDETR~\citep{mdetr2021}, GLIP~\citep{glip2022}, OWL-ViT~\citep{owlvit2022}, and Grounding DINO~\citep{groundingdino2024} support phrase or open-vocabulary localization, while Kosmos-2~\citep{kosmos22023}, Rex-Omni~\citep{rexomni2026}, and LocateAnything~\citep{locateanything2026} express boxes or points through generative vision-language interfaces. However, models trained mainly on natural images often fail to capture small, low-contrast, and morphologically subtle medical findings~\citep{medrpg2023,medsam2024}. Medical methods such as MedRPG~\citep{medrpg2023} and uMedGround~\citep{umedground2025} introduce domain-specific phrase and report grounding. MedROV~\citep{medrov2026} extends open-vocabulary detection to diverse medical imaging modalities, while VividMed~\citep{vividmed2024}, MIMO~\citep{mimo2025}, MedPLIB~\citep{medplib2024}, and UniBiomed~\citep{unibiomed2026} further support grounded biomedical interpretation through bounding boxes or segmentation masks. Nevertheless, existing medical grounding datasets and evaluations remain fragmented across imaging modalities, output formats, and task definitions~\citep{vividmed2024,biovil2022,medground2026}. This fragmentation limits joint training and consistent evaluation across heterogeneous medical images, motivating unified medical grounding data and systematic domain adaptation.

\textbf{Shared and Modality-Specific Representations in Medical Imaging.} Medical imaging modalities differ markedly in acquisition physics and visual appearance. Unified models such as MedSAM~\citep{medsam2024}, BiomedParse~\citep{biomedparse2025}, VividMed~\citep{vividmed2024}, and UniBiomed~\citep{unibiomed2026} learn shared representations from heterogeneous medical data, supporting multiple modalities and tasks within one framework. Full parameter sharing, however, may cause competing gradients and negative transfer when different modalities require distinct visual features~\citep{m4oe2024,unimed2024}. Recent MoE models introduce specialized computation to alleviate this problem. M4oE~\citep{m4oe2024} uses modality experts for medical image segmentation, MedMoE~\citep{medmoe2025} adapts visual representations to different imaging domains, and Uni-Med~\citep{unimed2024} employs connector experts to reduce multi-task interference. MedPLIB~\citep{medplib2024} further separates vision-language understanding from pixel-level grounding through expert modules. Most existing methods focus on segmentation, retrieval, VQA, or task-level routing. Modality specialization within a shared visual backbone for open-ended medical grounding remains underexplored, leaving the balance between shared knowledge and modality-specific representation unresolved.

\textbf{Reasoning-Augmented Visual Grounding.} Most grounding models predict spatial coordinates directly, providing limited insight into how a query is matched to visual evidence. Rex-Thinker~\citep{rexthinker2026} and UniVG-R1~\citep{univgr12025} address this limitation by introducing explicit chain-of-thought reasoning before object localization. In medicine, GEMeX~\citep{gemex2024} couples answers with textual and visual explanations, while MedCLM~\citep{medclm2025} constructs structured reasoning from lesion boxes and anatomical context. These studies suggest that intermediate rationales can make model predictions more traceable. However, longer reasoning does not guarantee faithful grounding. Medical chain-of-thought may amplify early perception errors when subtle abnormalities are not recognized correctly~\citep{bettereyes2026}. Reasoning supervision should therefore be built on reliable medical visual perception rather than used as a substitute for it. How to connect intermediate medical reasoning with the final spatial prediction remains an open problem.

\section{Method}

\begin{figure*}[t]
    \centering
    \includegraphics[
        width=\textwidth,
        trim={0.5cm 3.5cm 0.5cm 2.8cm},
        clip
    ]{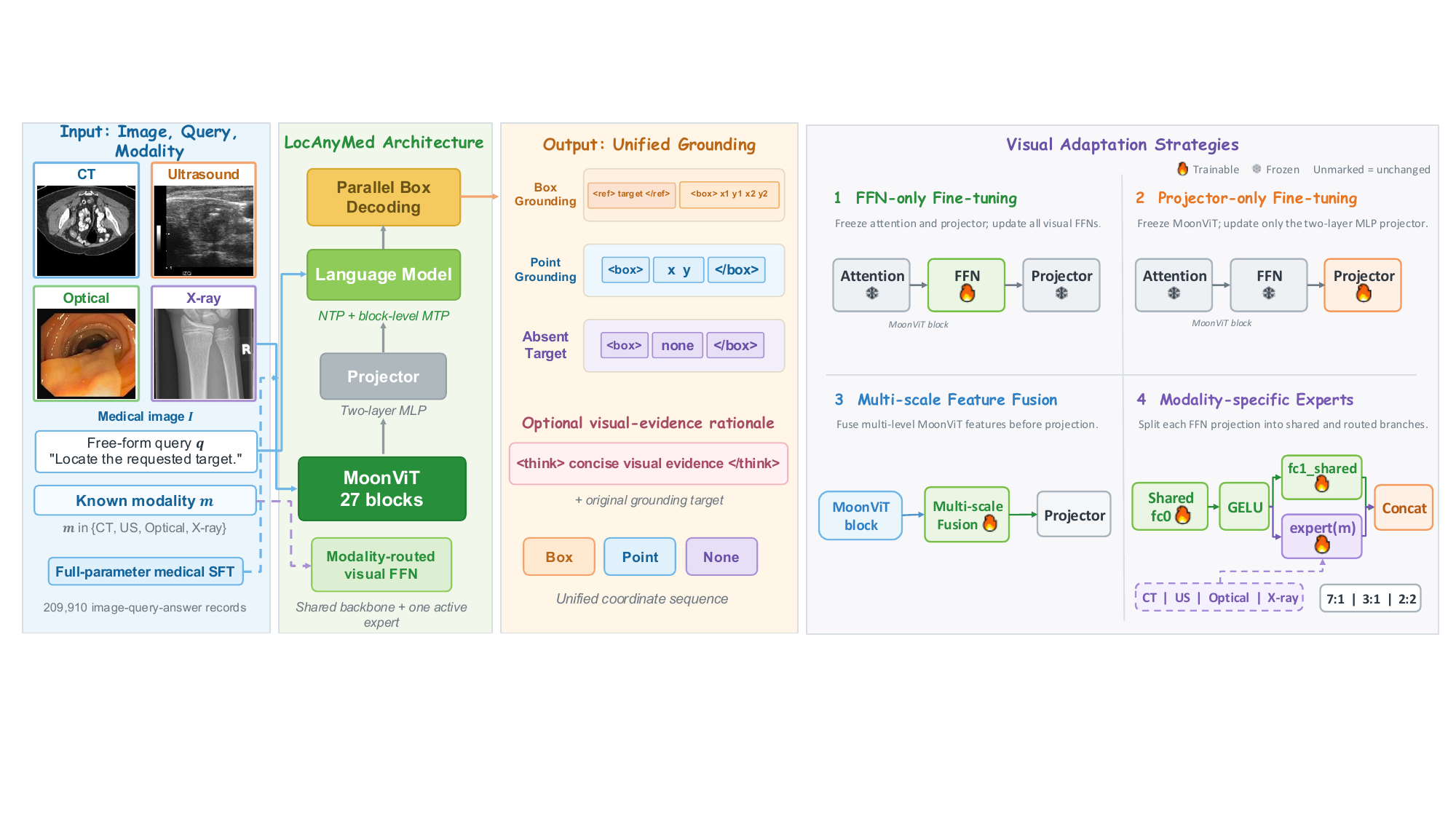}
    \caption{Overview of LocAnyMed and the evaluated visual adaptation strategies. \textbf{Left:} LocAnyMed takes a medical image, a free-form query, and its known imaging modality as input, and generates a unified box, point, or absent-target sequence. \textbf{Right:} the four evaluated strategies.}
    \label{Figure 2}
\end{figure*}

\subsection{Overview and Problem Formulation}

Given a medical image $I$, a free-form text query $q$, and its imaging modality $m$, medical visual grounding aims to generate the spatial location corresponding to the query. We consider four imaging modalities, with
$m \in \{\text{CT}, \text{Ultrasound}, \text{X-ray}, \text{Optical}\}$.
The supervision targets comprise bounding-box localization, point localization, and negative queries. Following LocateAnything~\cite{locateanything2026}, all targets are represented using a unified coordinate sequence: a bounding box is encoded by four coordinates, a point by two coordinates, and an absent target by the explicit token \verb|<box>none</box>|. Figure~\ref{Figure 2} summarizes LocAnyMed, which integrates unified medical grounding supervision, modality-routed visual adaptation, and visual-evidence rationale supervision within a shared vision--language architecture.

\subsection{Unified Medical Grounding Corpus Construction}

\begin{figure*}[t]
    \centering
    \includegraphics[
        width=\textwidth,
        trim={2cm 3.5cm 2cm 3cm},
        clip
    ]{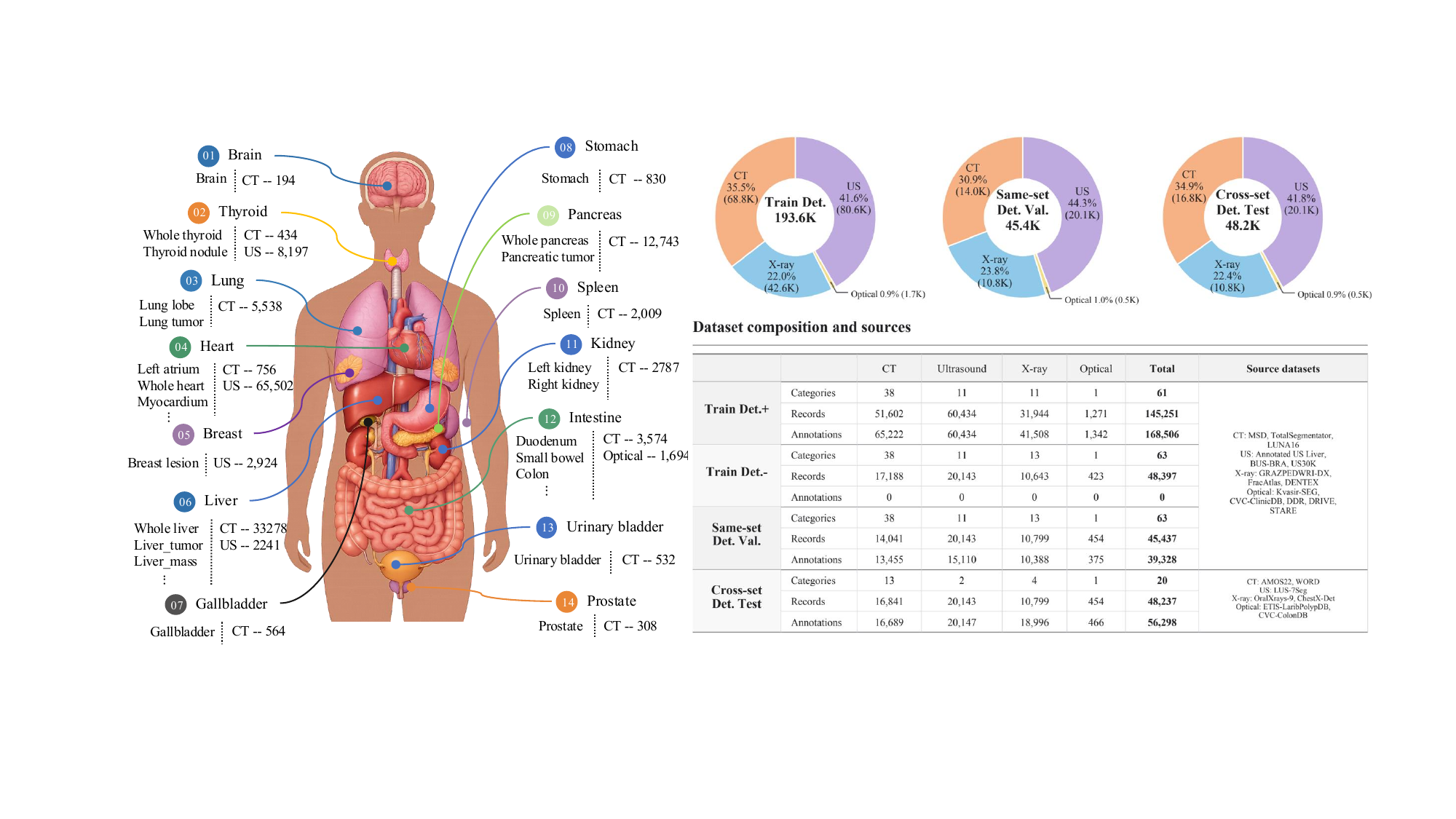}
    \caption{Composition of LocAnyMed-200K. \textbf{Left:} representative anatomical targets and record counts. \textbf{Right:} modality distributions, split statistics, and source datasets for Train Det.+, Train Det.-, Same-set Det. Val., and Cross-set Det. Test. Train Det.+ and Train Det.- denote target-present and absent-target detection records, respectively.}
    \label{Figure 3}
\end{figure*}

Existing medical datasets are typically organized by modality and annotation type, making unified visual grounding difficult. We therefore convert heterogeneous annotations into query-centered image–query–answer records, each containing an image $I$, query $q$, modality $m$, and serialized answer $a$. The corpus spans CT, ultrasound, X-ray, and optical imaging, with 193,648 detection records across 63 categories, including 145,251 target-present and 48,397 absent-target records.We refer to the union of Train Det.+ and Train Det.- as Train Det.; all specialized detector baselines are optimized on this same detection split, ensuring that their comparison with LocAnyMed uses identical medical training records rather than differences in data availability. An additional 16,262 point records are retained as auxiliary supervision, yielding 209,910 training records. Evaluation uses Same-set Det. Val. with 45,437 held-out records and Cross-set Det. Test with 48,237 records from datasets excluded from training. Figure~\ref{Figure 3} summarizes the anatomical targets, modality distributions, data splits, and sources.

The training corpus integrates medical datasets with different imaging mechanisms and annotation protocols. The CT subset is derived from the Medical Segmentation Decathlon~\citep{msd2022}, TotalSegmentator~\citep{totalsegmentator2023}, and LUNA16~\citep{luna162017}. The ultrasound subset combines US30K~\citep{samus2024}, BUS-BRA~\citep{busbra2024}, and Annotated Ultrasound. The X-ray subset is constructed from GRAZPEDWRI-DX~\citep{grazpedwri2022}, FracAtlas~\citep{fracatlas2023}, and DENTEX~\citep{dentex2023}. Positive polyp records in the optical subset are derived from Kvasir-SEG~\citep{kvasirseg2020} and CVC-ClinicDB~\citep{cvcclinicdb2015}. Additional optical source collections, including DDR~\citep{ddr2019}, DRIVE~\citep{drive2004}, and STARE~\citep{stare2000}, contribute images without polyp annotations for negative-query construction. We retain official dataset splits when available. Otherwise, source-derived records are grouped by case, patient, sequence, series, or image identifiers to prevent related positive annotations from appearing in both datasets.

We convert each source annotation according to its original spatial representation. For 2D or 3D segmentation masks, we perform eight-connected-component analysis on each 2D slice and use the minimum enclosing rectangle of each component as a bounding box. For tubular or thin structures in CT, the centroid of each connected component is used as a point target. World-coordinate annotations in LUNA16 are first transformed into voxel coordinates and then projected onto the corresponding axial slice. Bounding boxes provided in YOLO or COCO format are converted to pixel coordinates. All coordinates are subsequently discretized to $[0,1000]$ according to the actual width $W$ and height $H$ of the exported image:
\begin{equation}
\begin{aligned}
x_n &= \operatorname{round}
\left(\frac{x}{W}\times 1000\right),\\
y_n &= \operatorname{round}
\left(\frac{y}{H}\times 1000\right).
\end{aligned}
\end{equation}

Following the output representation of LocateAnything~\citep{locateanything2026}, a bounding box is serialized as \verb|<box><x1><y1><x2><y2></box>|, while a point reuses the two-coordinate format \verb|<box><x><y></box>|. A negative query is represented by the explicit null-localization token \verb|<box>none</box>|.Because the model resizes the entire image without spatial cropping, the normalized coordinates remain aligned with the visual input.

Queries are instantiated from deterministic templates, medical synonyms, and laterality expressions, whereas the \verb|<ref>| field retains the canonical category name. Target-present and absent-target detection records follow an approximately 3:1 ratio. Absent-target examples use standard localization queries without disclosing target absence and are paired with \verb|<box>none</box>|. Point grounding contains only target-present examples. Representative query--answer examples, together with source-specific conversion, template, and negative-sampling details, are provided in the supplementary material.

To complement coordinate-only grounding supervision with explicit visual evidence, we construct a rationale-augmented subset of exactly 20,000 detection records with unique image hashes. It contains 6,121 CT, 6,120 ultrasound, 6,120 X-ray, and 1,639 optical records, comprising 15,001 target-present and 4,999 absent-target examples; point records are excluded. We use GPT-5.5 to generate a concise visual-evidence rationale for each record. The teacher receives the complete medical image, target category, modality, and presence status. For target-present records, it additionally receives a target-region crop derived from the ground-truth box, but the prompt does not expose the original query or numerical coordinates. The generated rationale contains 20–45 words describing the visible appearance, relative location, surrounding structures, or evidence of absence. It is enclosed by \verb|<think>...</think>|, after which the original \verb|<ref>| and \verb|<box>| target is appended without modification. 

\subsection{LocAnyMed for Unified Medical Grounding Framework}

\begin{table*}[t]
\centering
\caption{
Overall box-grounding performance on Same-set Det. Val. and
Cross-set Det. Test. Specialized detectors are trained on Train Det.,
whereas general-purpose vision-language models and grounding specialists
are evaluated without medical fine-tuning. All values are percentages.
}
\label{tab:overall_results}

\footnotesize
\setlength{\tabcolsep}{3.5pt}
\renewcommand{\arraystretch}{0.98}

\begin{tabularx}{\textwidth}{
    @{}
    >{\raggedright\arraybackslash}p{0.29\textwidth}
    |CCC|CCC
    @{}
}
\toprule

\multirow{2}{*}{\textbf{Method}}
& \multicolumn{3}{c|}{
    \textbf{Same-set Det. Val.} (F1@IoU, \%)
  }
& \multicolumn{3}{c}{
    \textbf{Cross-set Det. Test} (F1@IoU, \%)
  } \\

\cmidrule(lr){2-4}
\cmidrule(lr){5-7}

& \textbf{0.50}
& \textbf{0.95}
& \textbf{Mean}
& \textbf{0.50}
& \textbf{0.95}
& \textbf{Mean} \\
\midrule

\multicolumn{7}{c}{\textbf{Open-set Specialized Detectors}} \\
\midrule

Grounding DINO
& 51.21 & 11.82 & 40.35
& 20.09 & 1.28 & 11.33 \\

\midrule
\multicolumn{7}{c}{\textbf{Closed-set Specialized Detectors}} \\
\midrule

DETR-R50
& 45.73 & 10.98 & 35.75
& 12.91 & 0.43 & 6.88 \\

EdgeCrafter
& 79.98 & 8.42 & 61.00
& 1.94 & 0.25 & 1.43 \\

YOLO26
& 78.61 & 16.08 & 63.72
& 0.77 & 0.12 & 0.62 \\

LW-DETR-DINOv2
& 81.34 & 17.00 & 64.78
& 1.32 & 0.20 & 0.99 \\

LW-DETR-Objects365
& 82.58 & 18.71 & 66.25
& 1.41 & 0.21 & 1.08 \\

\midrule
\multicolumn{7}{c}{\textbf{Vision-Language Models}} \\
\midrule

DeepSeek-VL2-Small
& 7.87 & 0.44 & 3.60
& 8.87 & 0.08 & 2.78 \\

MiMo-VL-7B
& 10.84 & 0.18 & 4.00
& 13.76 & 0.01 & 3.57 \\

LocateAnything-3B
& 10.64 & 0.59 & 5.72
& 11.87 & 0.42 & 5.70 \\

Ovis2.5-2B
& 18.13 & 0.03 & 5.90
& 9.08 & 0.00 & 2.21 \\

Rex-Omni-3B
& 15.48 & 0.69 & 9.77
& 10.39 & 0.11 & 4.97 \\

GPT-5.5
& 30.24 & 0.24 & 13.74
& 29.07 & 0.05 & 11.84 \\

Qwen3-VL-4B
& 29.82 & 9.70 & 18.36
& 25.89 & 5.24 & 12.17 \\

Qwen3-VL-32B
& 29.91 & 17.61 & 23.27
& 27.49 & \textbf{9.28} & 16.02 \\

Qwen3-VL-8B
& 36.03 & 18.60 & 25.91
& 26.42 & 8.48 & 14.86 \\

\midrule
\textbf{LocAnyMed}
& \textbf{85.59}
& \textbf{37.43}
& \textbf{69.15}
& \textbf{33.63}
& 1.58
& \textbf{19.12} \\

\bottomrule
\end{tabularx}
\end{table*}

LocAnyMed extends the open-ended grounding capability of LocateAnything~\citep{locateanything2026} to heterogeneous medical images. We retain its overall vision–language architecture and unified grounding decoder, and introduce modality specialization into the output projection of each visual FFN. MoonViT contains 27 Transformer blocks. Omitting the residual connection, the original dense FFN in each block is formulated as
\begin{equation}
\begin{aligned}
h=\operatorname{GELU}(W_0x+b_0),\qquad y=W_1h+b_1,
\end{aligned}
\end{equation}
where $W_0\in\mathbb{R}^{4304\times1152}$ expands the input features and $W_1\in\mathbb{R}^{1152\times4304}$ projects them back to the original dimension. In the dense architecture, CT, ultrasound, X-ray, and optical images share the entire FFN. A single shared projection provides no explicit capacity for modality-specific visual transformations. We therefore replace $W_1$ in all 27 visual blocks with a modality-routed shared-specific projection. All remaining model components are structurally unchanged, shared across modalities, and jointly optimized during full-parameter SFT. Figure~\ref{Figure 3}  summarizes the overall framework and the modality-routed visual FFN.

As illustrated in Figure~\ref{Figure 3}, let $h=\operatorname{GELU}(W_0x+b_0)$ denote the output of the shared expansion layer. We decompose the original projection $W_1$ into a shared branch $W_s$ and four modality-specific branches $\{W_m\}_{m=1}^{4}$. For a visual token assigned to modality $m$, the FFN output is

\begin{equation}
\begin{aligned}
y=\operatorname{Concat} \left( W_sh+b_s,\; W_mh+b_m \right),
\end{aligned}
\end{equation}
where $W_s\in\mathbb{R}^{d_s\times4304}$, $W_m\in\mathbb{R}^{d_e\times4304}$, and $d_s+d_e=1152$. The shared branch is activated for every input, whereas $W_m$ is selected using the known CT, X-ray, ultrasound, or optical modality. For a packed sequence containing samples from multiple modalities, each image-level modality label is expanded to its corresponding visual tokens, allowing different tokens to enter their respective experts. Routing is deterministic and introduces no learned router, gating weight, or load-balancing loss. A valid modality label is therefore required during both training and inference.

We vary the channel allocation between the shared and modality-specific branches to study their capacity trade-off. The 7:1, 3:1, and 2:2 settings use $(d_s,d_e)=(1008,144)$, $(864,288)$, and $(576,576)$, respectively, with $d_s+d_e=1152$ in all cases. To preserve the pretrained visual representation, each branch is initialized from the original dense projection $W_1,b_1$:

\begin{equation}
\begin{aligned}
W_s &= W_1[0:d_s,:], \quad b_s=b_1[0:d_s],\\
W_m &= W_1[d_s:1152,:], \quad b_m=b_1[d_s:1152],
\quad \forall m .
\end{aligned}
\end{equation}

The modality-specific slice is copied to all four experts, so they start from identical parameters. Consequently, concatenating the shared output with any selected expert recovers the original dense projection at initialization. During training, the shared branch provides capacity for cross-modality feature sharing, while each expert can learn modality-specific transformations. Because only one expert is activated per input, the active output width and nominal multiply-add count of the projection remain unchanged. 

\subsection{Training Objective and Model Variants}
Following LocateAnything~\citep{locateanything2026}, we supervise each assistant response using both its autoregressive NTP sequence and an appended block-level MTP sequence. Let $\mathcal I_{\mathrm{NTP}}$ and $\mathcal I_{\mathrm{MTP}}$ denote their valid label positions. The model is optimized using a single mean cross-entropy objective:
\begin{equation}
\begin{aligned}
\mathcal{L}
&=
\frac{
\sum\limits_{i\in\mathcal I_{\mathrm{NTP}}}\ell_i
+
\sum\limits_{j\in\mathcal I_{\mathrm{MTP}}}\ell_j
}
{
N_{\mathrm{NTP}}+N_{\mathrm{MTP}}
}.
\end{aligned}
\end{equation}

No explicit coefficient is used to balance the two forms; their relative contributions are determined by the numbers of valid tokens. Image and query tokens are excluded as prediction targets, while the assistant loss still propagates through the visual features to the encoder. Box, point, negative, and rationale tokens share the same objective, without task-specific or routing losses.

\begin{table}[t]
\centering
\caption{
Ablation of visual adaptation strategies, shared-to-expert channel allocation,
and visual-evidence rationale supervision. For the projector-only,
visual-fc1-only, and multi-scale variants, we report the checkpoint selected
by Same-set F1 Mean, with F1@IoU 0.50 used to break ties.
All results are percentages.
}
\label{tab:ablation}
\scriptsize
\setlength{\tabcolsep}{3.2pt}
\renewcommand{\arraystretch}{0.94}

\begin{tabular*}{\columnwidth}{@{\extracolsep{\fill}}lcccc@{}}
\toprule
\multirow{2}{*}{\textbf{Adaptation}}
& \multicolumn{2}{c}{\textbf{Same-set Det. Val.}}
& \multicolumn{2}{c}{\textbf{Cross-set Det. Test}} \\
\cmidrule(lr){2-3}\cmidrule(lr){4-5}
& \textbf{F1@0.50} & \textbf{F1 Mean}
& \textbf{F1@0.50} & \textbf{F1 Mean} \\
\midrule

\multicolumn{5}{c}{\textit{Parameter-efficient adaptation}} \\
\midrule
Projector only
& 61.48 & 42.18 & 25.79 & 12.62 \\
Visual FFN fc1 only
& 76.92 & 56.88 & 19.65 & 10.47 \\

\midrule
\multicolumn{5}{c}{\textit{Full-parameter adaptation}} \\
\midrule
LocAnyMed
& \underline{85.59} & \underline{69.15}
& 33.63 & 19.12 \\
LocAnyMed + Multi-scale
& 84.77 & 66.52
& \underline{35.11} & \underline{20.97} \\

\midrule
\multicolumn{5}{c}{\textit{Modality-routed experts}} \\
\midrule
LocAnyMed + MoE (3:1)
& \textbf{85.73} & \textbf{69.17}
& 31.68 & 18.39 \\
LocAnyMed + MoE (2:2)
& 85.48 & 68.87
& 33.90 & 19.51 \\
LocAnyMed + MoE (7:1)
& 85.52 & 69.02
& 32.42 & 18.39 \\

\midrule
\multicolumn{5}{c}{\textit{Visual-evidence rationale supervision}} \\
\midrule
LocAnyMed + MoE (2:2) + Rationale
& 85.30 & 68.96
& \textbf{35.49} & \textbf{21.34} \\
LocAnyMed + Rationale
& 85.38 & 69.05
& 34.49 & 20.10\\

\bottomrule
\end{tabular*}
\end{table}

\begin{figure*}[t]
    \centering
    \includegraphics[
        width=\textwidth,
        trim={1.8cm 6.5cm 1.8cm 2.5cm},
        clip
    ]{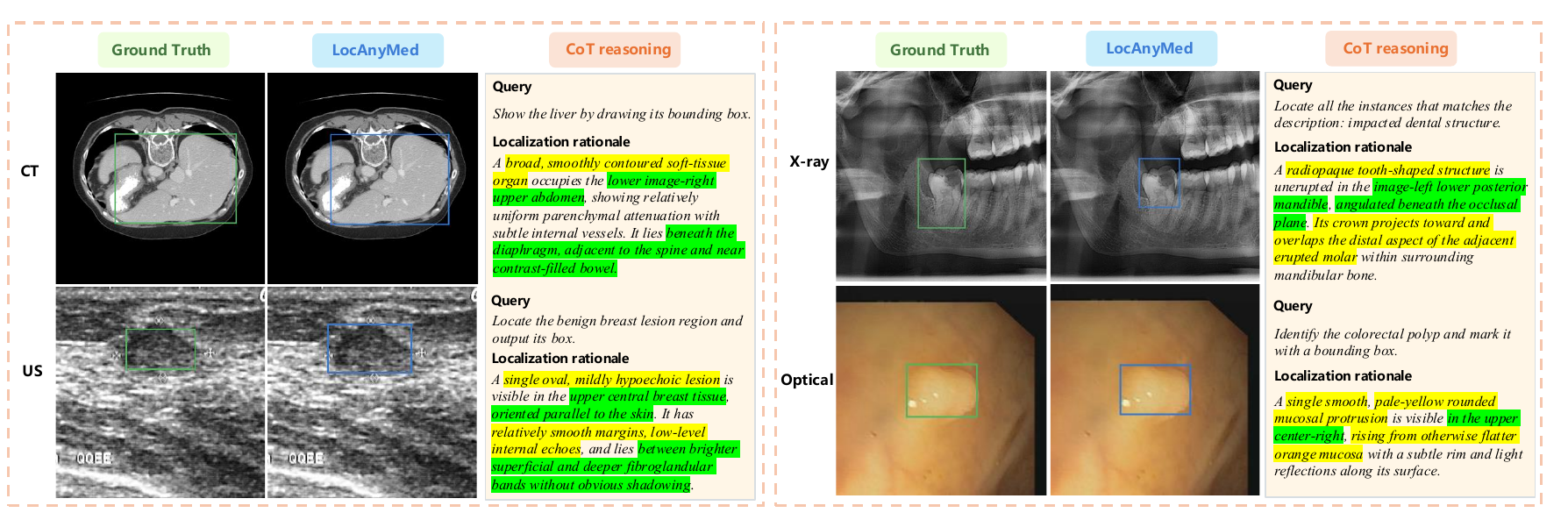}
    \caption{
    Qualitative examples of rationale-augmented grounding across CT, ultrasound, X-ray, and optical imaging. Ground-truth and predicted boxes are shown in green and blue, respectively. Each example includes the input query and the generated visual-evidence rationale. Yellow highlights denote descriptions of target appearance, while green highlights denote spatial
    location and anatomical relationships.
    }
    \label{Figure 5}
\end{figure*}

\section{Experiments}

\subsection{Training Details and Evaluation Setup}
\textbf{Training Details.} We first perform full-parameter SFT of LocateAnything-3B~\citep{locateanything2026} on the unified medical grounding corpus of 209,910 records; the resulting model is denoted as LocAnyMed. To introduce modality-specific visual capacity, we construct 7:1, 3:1, and 2:2 MoE configurations from the same pretrained LocateAnything-3B checkpoint and train each configuration independently on the same medical corpus. The selected 2:2 MoE checkpoint is subsequently fine-tuned on the 20,000-record visual-evidence rationale corpus. The rationale stage changes the supervision sequence without modifying the model architecture. All visual, vision–language connection, and language-model parameters remain trainable throughout training.

\textbf{Compared Methods.} We compare LocAnyMed with three groups of methods: specialized detectors, including Grounding DINO~\citep{groundingdino2024}, DETR-R50~\citep{detr2020}, YOLO26~\citep{yolo262026}, LW-DETR~\citep{lwdetr2024}, and EdgeCrafter~\citep{edgecrafter2026}; general-purpose VLMs, including DeepSeek-VL2-Small~\citep{deepseekvl22024}, MiMo-VL-7B~\citep{mimovl2025}, Ovis2.5-2B~\citep{ovis252025}, Qwen3-VL~\citep{qwen3vl2025}, and GPT-5.5; and grounding specialists, including Rex-Omni-3B~\citep{rexomni2026} and LocateAnything-3B~\citep{locateanything2026}. Specialized detectors are trained on Train Det., whereas the VLMs and grounding specialists are evaluated without medical fine-tuning.

\textbf{Evaluation Setup.} We report F1@IoU at thresholds of 0.50 and 0.95, together with F1 Mean averaged over thresholds from 0.50 to 0.95 in increments of 0.05. Scores are computed from dataset-level true positives, false positives, and false negatives under one-to-one box matching. All methods are evaluated on both datasets.

\subsection{Overall Medical Grounding Results}
We evaluate box grounding on Same-set Det. Val. and Cross-set Det. Test. Table~\ref{tab:overall_results} presents the overall comparison, while Table~\ref{tab:ablation} reports the visual adaptation and rationale-supervision ablations. Detailed modality-wise results, point localization, throughput, and latency are provided in the supplementary material.

\textbf{Overall Comparison.} Table~\ref{tab:overall_results} reports the overall results. On Same-set Det. Val., LocAnyMed achieves 85.59 F1@IoU 0.50 and 69.15 F1 Mean, raising the F1 Mean of LocateAnything-3B from 5.72 to 69.15. It also exceeds LW-DETR-Objects365 by 2.90 F1 Mean and 18.72 F1@IoU 0.95. Under cross-set evaluation, the F1 Mean of LW-DETR-Objects365 and EdgeCrafter falls to 1.08 and 1.43, respectively, whereas LocAnyMed retains an F1@IoU 0.50 of 33.63 and an F1 Mean of 19.12. The corresponding F1 Mean of LocateAnything-3B is 5.70, demonstrating the importance of unified medical grounding supervision for cross-source generalization.

\textbf{Rationale-Augmented LocAnyMed.} As shown in Table~\ref{tab:ablation}, rationale supervision improves the Cross-set F1@IoU 0.50 from 33.90 to 35.49 and the F1 Mean from 19.51 to 21.34. Same-set performance remains stable: F1@IoU 0.50 changes slightly from 85.48 to 85.30, while F1 Mean increases from 68.87 to 68.96. This pattern suggests that describing target appearance, relative location, and surrounding anatomy encourages the model to use medically relevant visual evidence. Meanwhile, retaining the original grounding target after each rationale preserves direct localization supervision. Rationale supervision therefore improves cross-source robustness without compromising overall same-set localization. Figure~\ref{Figure 5} presents representative predictions and generated rationales.

\subsection{Ablation Studies}
\textbf{Visual Adaptation Strategies.} Table~\ref{tab:ablation} compares parameter-restricted and full-parameter visual adaptation strategies. Fine-tuning only the projector achieves a Same-set F1 Mean of 42.18, while adapting the output projection of each visual FFN increases it to 56.88. Both strategies remain substantially below full-parameter LocAnyMed, which achieves a Same-set F1@IoU 0.50 of 85.59 and an F1 Mean of 69.15. Multi-scale fusion obtains comparable Same-set performance and improves Cross-set F1@IoU 0.50 and F1 Mean from 33.63 and 19.12 to 35.11 and 20.97, respectively. These results indicate that restricted parameter updates are insufficient for medical grounding, whereas aggregating features from the last three MoonViT blocks improves robustness across data sources.

\textbf{Expert Allocation and Rationale Supervision.} Different shared-to-expert allocations favor different evaluation settings. The 3:1 configuration achieves the best Same-set results, with an F1@IoU 0.50 of 85.73 and an F1 Mean of 69.17. In contrast, the 2:2 configuration performs best among the expert variants on Cross-set Det. Test, reaching 33.90 and 19.51, respectively. We therefore select the 2:2 allocation for subsequent rationale-supervised fine-tuning because of its stronger cross-source performance. Adding visual-evidence rationale supervision further increases Cross-set F1@IoU 0.50 from 33.90 to 35.49 and F1 Mean from 19.51 to 21.34. This result suggests that rationale supervision improves cross-source robustness.

\section{Conclusion}
We present LocAnyMed, a unified grounding framework for CT, ultrasound, X-ray, and optical medical images. Full-parameter SFT on 209,910 unified records substantially improves the medical grounding capability and cross-source generalization of LocateAnything-3B. Comparisons with specialized detectors, general-purpose vision--language models, and grounding specialists further demonstrate the importance of large-scale domain-specific supervision. Visual-evidence rationale supervision encourages the model to associate target appearance with anatomical and spatial context, improving cross-set robustness while preserving same-set performance. Together, LocAnyMed-200K and LocAnyMed-CoT-20K provide a foundation for studying accurate and inspectable grounding across heterogeneous medical images. Future work will expand the dataset with additional sources, modalities, and target categories; combine chain-of-thought supervision with reinforcement learning to improve generalization; and distill LocAnyMed into a compact model
for efficient deployment.

\clearpage
\appendix
\section*{Supplementary Material}
\setcounter{secnumdepth}{2}
\setcounter{section}{0}
\setcounter{table}{0}
\setcounter{figure}{0}
\setcounter{equation}{0}
\renewcommand{\thesection}{S\arabic{section}}
\renewcommand{\thesubsection}{\thesection.\arabic{subsection}}
\renewcommand{\thetable}{S\arabic{table}}
\renewcommand{\thefigure}{S\arabic{figure}}
\renewcommand{\theequation}{S\arabic{equation}}
\renewcommand{\theHsection}{supp.\arabic{section}}
\renewcommand{\theHsubsection}{supp.\arabic{section}.\arabic{subsection}}
\renewcommand{\theHtable}{supp.table.\arabic{table}}
\renewcommand{\theHfigure}{supp.figure.\arabic{figure}}
\renewcommand{\theHequation}{supp.equation.\arabic{equation}}
\setlength{\textfloatsep}{10pt plus 2pt minus 2pt}
\setlength{\floatsep}{8pt plus 2pt minus 2pt}
\setlength{\intextsep}{8pt plus 2pt minus 2pt}
\renewcommand{\topfraction}{0.92}
\renewcommand{\bottomfraction}{0.80}
\renewcommand{\textfraction}{0.05}
\renewcommand{\floatpagefraction}{0.75}
\setcounter{topnumber}{3}
\setcounter{bottomnumber}{2}
\setcounter{totalnumber}{5}

\section{Dataset Construction and Serialization}

\subsection{Record Definition}
LocAnyMed-200K contains image-query-answer records, not 209,910 independent images. Multiple queries may refer to the same underlying image, but all records derived from a shared case, patient, sequence, series, or image identifier are kept in one split whenever that identifier is available. Table~\ref{supp:tab:train-composition} gives the exact training composition. The 193,648 detection records include all 48,397 absent-target queries; they are not an additional set.

\subsection{Data Sources and Splits}

Table~\ref{supp:tab:sources} lists the datasets used for training and same-set validation, together with the independent datasets reserved for cross-set evaluation. Data assigned to the \cross{} split are excluded from
fine-tuning.

For CT, Medical Segmentation Decathlon and TotalSegmentator provide 3-D semantic masks. In each exported 2-D slice, eight-connected components are converted into bounding boxes for compact structures, whereas component
centroids are used for thin or tubular structures. LUNA16 nodule coordinates are projected from world coordinates into voxel space to construct point targets. Official case splits are retained when available; otherwise, a deterministic 80/20 split is performed at the case or series level. For ultrasound, processed box annotations derived from segmentation masks or detector labels are mapped to canonical category names. The original box task is preserved, and box-center points are not synthesized. Existing train/validation/test groupings are retained, with validation and test partitions combined only when this does not cross source-defined groups. For X-ray, native boxes, COCO-style annotations, and fracture masks are converted into the common box-coordinate representation. No box-center points are generated. GRAZPEDWRI-DX is partitioned at the patient level, FracAtlas uses its official split, and datasets without patient identifiers use a deterministic image-level split. For the optical modality, Kvasir-SEG provides native polyp boxes, while connected components from CVC-ClinicDB masks are converted into enclosing boxes. Kvasir-SEG is split deterministically at the image level, whereas CVC-ClinicDB is partitioned at the sequence level. DDR, DRIVE, and STARE are used only to construct polyp-absent negative queries. Their retinal vessel masks and image-level disease labels are not converted into polyp boxes.

All \cross{} datasets use the same serialization and coordinate conversion as the training data. To prevent source leakage, they are checked against the training and \same{} partitions using source names, file basenames, and
perceptual hashes.

\begin{table}[!t]
\centering
\caption{Exact LocAnyMed-200K training-record composition.  ``Det.+'' and
``Det.-'' denote target-present and absent-target detection queries.}
\label{supp:tab:train-composition}
\footnotesize
\setlength{\tabcolsep}{5pt}
\renewcommand{\arraystretch}{0.95}
\begin{tabular}{@{}lrrrr@{}}
\toprule
Modality & Det.+ & Det.- & Point & Total \\
\midrule
CT         & 51,602 & 17,188 & 16,262 & 85,052 \\
Ultrasound & 60,434 & 20,143 & 0      & 80,577 \\
X-ray      & 31,944 & 10,643 & 0      & 42,587 \\
Optical    & 1,271  & 423    & 0      & 1,694 \\
\midrule
Total      & 145,251 & 48,397 & 16,262 & 209,910 \\
\bottomrule
\end{tabular}
\end{table}

\begin{table}[t]
\centering
\caption{Datasets used for training and same-set validation, and for
cross-set evaluation.}
\label{supp:tab:sources}
\small

\begin{tabularx}{\columnwidth}{
  p{0.20\columnwidth}
  >{\raggedright\arraybackslash}X
}
\toprule
Modality & Sources by split \\
\midrule

CT
&
\textbf{Training/\same{}:}
Medical Segmentation Decathlon, TotalSegmentator, and LUNA16.
\newline
\textbf{\cross{}:}
AMOS22 and WORD.
\\

\addlinespace[2pt]

Ultrasound
&
\textbf{Training/\same{}:}
US30K collections, BUS-BRA, and Annotated Ultrasound Liver.
\newline
\textbf{\cross{}:}
LUS-7Seg.
\\

\addlinespace[2pt]

X-ray
&
\textbf{Training/\same{}:}
GRAZPEDWRI-DX, FracAtlas, and DENTEX.
\newline
\textbf{\cross{}:}
OralXrays-9 and ChestX-Det.
\\

\addlinespace[2pt]

Optical
&
\textbf{Training/\same{}:}
Kvasir-SEG and CVC-ClinicDB; DDR, DRIVE, and STARE are used only for
negative-query construction.
\newline
\textbf{\cross{}:}
ETIS-LaribPolypDB and CVC-ColonDB.
\\

\bottomrule
\end{tabularx}
\end{table}

\subsection{Coordinates, targets, and language}
For an exported image of width $W$ and height $H$, pixel coordinates are discretized as
\begin{equation}
\begin{aligned}
  x_n &= \operatorname{round}\left(\frac{1000x}{W}\right), \\
  y_n &= \operatorname{round}\left(\frac{1000y}{H}\right).
\end{aligned}
\end{equation}
The complete image is resized by the model processor without spatial cropping or padding that changes this coordinate frame. Compact mask components yield minimum enclosing boxes.  Thin or tubular CT structures yield component centroids. LUNA16 world coordinates are transformed using origin, spacing, and direction, then projected onto the corresponding axial slice. Points are therefore derived only from native point annotations or appropriate mask centroids, never from box centers.

Detection queries are deterministically sampled from 50 box-eliciting templates. Point queries use a disjoint five-template pool. Target-specific synonyms, natural anatomical names, and laterality expressions may appear in the query, while the answer retains a canonical reference name. For each detection target and split, absent-target records are sampled to approximately a 3:1 positive-to-negative ratio. The negative query uses the same ordinary localization templates and does not reveal that the target is absent.

\begin{center}
\scriptsize
\setlength{\tabcolsep}{3pt}
\renewcommand{\arraystretch}{0.92}

\begin{tabularx}{\columnwidth}{p{0.27\columnwidth}Y}
\multicolumn{2}{p{\columnwidth}}{\textbf{Detection-query examples.}
Template examples show the deterministic pool before target substitution; generated examples show actual medical queries after synonym and laterality
insertion.} \\
\addlinespace[2pt]
\toprule
Type & Examples \\
\midrule
Templates &
\texttt{Find all \{target\} and give me their bounding boxes.};
\texttt{Please outline the \{target\} with a bounding rectangle.};
\texttt{Locate all instances of \{target\}.};
\texttt{Where is the \{target\}? Provide the bounding box.};
\texttt{Identify every visible \{target\} in the image.} \\
\midrule
Generated queries &
\texttt{Find all left suprarenal gland and give me their bounding boxes.};
\texttt{Please outline the left suprarenal gland with a bounding rectangle.};
\texttt{Locate all colon polyps.};
\texttt{Identify every visible pulmonary nodule in the image.};
\texttt{Where is the aorta?} \\
\midrule
Absent-target query &
The same template pool is used, e.g.,
\texttt{Please outline the left suprarenal gland with a bounding rectangle.}
may be paired with \texttt{<box>none</box>} when that target is absent. \\
\bottomrule
\end{tabularx}
\end{center}

\section{Additional Implementation Details}

\subsection{Modality-routed visual FFNs}
MoonViT has 27 Transformer blocks with dense FFNs $1152\rightarrow4304\rightarrow1152$.  In every block, only the output projection $W_1\in\mathbb{R}^{1152\times4304}$ is decomposed. The input projection, GELU, LayerNorm, attention, vision--language projector, and language model remain structurally shared and are updated during full-parameter SFT.

There are exactly four experts, ordered CT, X-ray, ultrasound, and optical. For a shared width $d_s$ and active expert width $d_e$, $d_s+d_e=1152$, and
\begin{equation}
  y=\operatorname{Concat}[W_s h+b_s,\;W_m h+b_m].
\end{equation}
The 7:1, 3:1, and 2:2 allocations use $(d_s,d_e)=(1008,144)$, $(864,288)$, and $(576,576)$.  $W_s,b_s$ copy the first $d_s$ output channels of the dense projection.  Every expert copies the remaining $d_e$ channels. Thus, concatenation exactly reproduces the original dense projection at initialization for every route.

Routing reads the dataset modality metadata; it has no learned gate or load-balancing loss.  Mixed-modality packed batches are supported because each image label is expanded to its visual tokens before dispatch.  A missing, unknown, or length-mismatched modality label raises an error rather than silently selecting a default expert.  Inference therefore requires the known imaging modality.

\subsection{Multi-scale fusion}
We evaluate two placements for the same top-down fusion rule.  The \emph{last-three} strategy captures MoonViT blocks 25, 26, and 27, treating the final three visual layers as a local refinement hierarchy. Features from blocks 25 and 26 are adaptively average-pooled by factors four and two.  Let $C,M,F$ denote the coarse, medium, and fine feature grids.  Bilinear top-down fusion is
\begin{align}
  P_M &= \tfrac12\left(M+\operatorname{Up}(C)\right),\\
  P_F &= \tfrac12\left(F+\operatorname{Up}(P_M)\right),\\
  Y   &= \tfrac12(F+P_F).
\end{align}
The averages prevent an uncalibrated increase in activation scale. The \emph{grouped-stage} strategy instead divides the 27 MoonViT blocks into three groups, takes the outputs of blocks 9, 18, and 27, applies the same four-times and two-times pooling to the first two group outputs, and fuses
them into the block-27 representation with the same averaging rule. Thus the two variants differ only in where multi-scale features are tapped: local late-block refinement for last-three fusion versus wider stage-separated aggregation for grouped-stage fusion.

\subsection{Loss and optimization}
LocateAnything's dual supervision is retained.  Each assistant answer supplies an autoregressive next-token sequence and an appended block-level MTP sequence. With valid positions $\mathcal I_{\mathrm{NTP}}$ and
$\mathcal I_{\mathrm{MTP}}$, the implementation minimizes
\begin{equation}
\mathcal L=\frac{\sum_{i\in\mathcal I_{\mathrm{NTP}}}\ell_i+
\sum_{j\in\mathcal I_{\mathrm{MTP}}}\ell_j}
{N_{\mathrm{NTP}}+N_{\mathrm{MTP}}}.
\end{equation}
There is no separate coefficient, detection loss, point loss, rationale loss, or routing loss.  Image and human-query tokens are masked as prediction targets.  All assistant tokens---including \texttt{<think>}, \texttt{<ref>}, box coordinates, points, and \texttt{<box>none</box>}---participate in the same token objective.

The main 200K runs use four H20 GPUs, bfloat16 training, per-device batch size one, gradient accumulation eight, and an effective packed-sequence batch size of 32.  We use learning rate $2\times10^{-5}$, image-token limit 1024, sequence/token limit 8192, block size six, Magi language attention, FlashAttention-2 for vision, gradient checkpointing, and DeepSpeed ZeRO-2. The training loader uses stream-packed MTP sequences: one optimizer update is a packed-sequence update, and the number of raw image--query--answer records inside a packed sequence varies with sequence length.  We therefore compare checkpoints by optimization step rather than by exact epoch.  CoT continuation uses learning rate $5\times10^{-6}$, a linear schedule, no warm-up, and checkpoints every 100 steps under the same packed-loader convention.

\section{Evaluation}

\subsection{Frozen evaluation sets}
Table~\ref{supp:tab:eval-composition} gives the exact evaluation-set composition. Same-set Det. Val. contains 45,437 detection records, whereas Cross-set Det. Test contains 48,237 records. The two sets are not paired at the image level. \cross{} contains 20 modality--category pairs, all of which are also represented in training, but every image source is excluded from training. The \same{} common-category control contains 13,379 records restricted to those same 20 pairs; it separates source shift from a simple category-set mismatch. In compact result tables below, ``Same'' denotes \same{} and ``Cross'' denotes \cross{}; these are abbreviations of the paper's two evaluation splits, not additional dataset names.

\begin{table}[t]
\centering
\caption{Frozen evaluation records by modality.}
\label{supp:tab:eval-composition}
\footnotesize
\setlength{\tabcolsep}{5pt}
\renewcommand{\arraystretch}{0.95}
\begin{tabular}{@{}lrrr@{}}
\toprule
Modality & Same det. & Same point & Cross det. \\
\midrule
CT         & 14,041 & 2,800 & 16,841 \\
Ultrasound & 20,143 & 0     & 20,143 \\
X-ray      & 10,799 & 0     & 10,799 \\
Optical    & 454    & 0     & 454 \\
\midrule
Total      & 45,437 & 2,800 & 48,237 \\
\bottomrule
\end{tabular}
\end{table}

\subsection{Matching and metrics}
For each query independently, predicted and ground-truth boxes form an IoU matrix.  At each threshold $t\in\{0.50,0.55,\ldots,0.95\}$, a one-to-one maximum-cardinality bipartite matching is computed over edges with IoU at
least $t$.  Dataset-level TP, FP, and FN are then summed before computing precision, recall, and F1.  F1 Mean is the arithmetic mean of the ten F1 values.  There is no confidence threshold: the generative model emits only coordinates or an explicit null response.  Macro-F1 first computes these metrics for every modality/category pair and then averages categories equally.

An absent-target prediction is correct only when the parsed output is exactly an explicit null localization; malformed or truncated outputs are incorrect. Point predictions use one-to-one matching under Euclidean distance normalized by the image diagonal.  PCK@0.05 is the primary point-localization metric; we also report PCK@0.10 and nearest normalized distance as looser and continuous diagnostics.

Evaluation runs one model replica per H20 GPU with four data shards.  Inference uses float16, SDPA, batch size one, deterministic generation (\texttt{do\_sample=False}, temperature zero), original per-record natural language queries, image-token limit 1024, and the model's slow generation path. Detection and point limits are 2048 and 512 new tokens, with an 8192-token retry only upon truncation.  CoT models append the trained \texttt{/think} trigger to detection prompts; point prompts remain unchanged because CoT-20K contains no point records.

\section{Unified Medical Adaptation}

\begin{table*}[t]
\centering
\caption{Modality-wise box-grounding performance on Same-set Det. Val. and Cross-set Det. Test. All values are percentages. F1 Mean is averaged over IoU thresholds from 0.50 to 0.95 with a step size of 0.05.}
\label{supp:tab:modality_results}
\scriptsize
\setlength{\tabcolsep}{2.2pt}
\renewcommand{\arraystretch}{1.05}

\resizebox{\textwidth}{!}{%
\begin{tabular}{@{}l|cc|cc|cc|cc||cc|cc|cc|cc@{}}
\toprule

\multirow{3}{*}{\textbf{Method}}
& \multicolumn{8}{c||}{
  \textbf{Same-set Det. Val.} (F1@IoU, \%)
}
& \multicolumn{8}{c}{
  \textbf{Cross-set Det. Test} (F1@IoU, \%)
} \\

\cmidrule(lr){2-9}
\cmidrule(lr){10-17}

& \multicolumn{2}{c|}{\textbf{CT}}
& \multicolumn{2}{c|}{\textbf{Ultrasound}}
& \multicolumn{2}{c|}{\textbf{Optical}}
& \multicolumn{2}{c||}{\textbf{X-ray}}
& \multicolumn{2}{c|}{\textbf{CT}}
& \multicolumn{2}{c|}{\textbf{Ultrasound}}
& \multicolumn{2}{c|}{\textbf{Optical}}
& \multicolumn{2}{c}{\textbf{X-ray}} \\

\cmidrule(lr){2-3}
\cmidrule(lr){4-5}
\cmidrule(lr){6-7}
\cmidrule(lr){8-9}
\cmidrule(lr){10-11}
\cmidrule(lr){12-13}
\cmidrule(lr){14-15}
\cmidrule(lr){16-17}

& \textbf{0.50} & \textbf{Mean}
& \textbf{0.50} & \textbf{Mean}
& \textbf{0.50} & \textbf{Mean}
& \textbf{0.50} & \textbf{Mean}
& \textbf{0.50} & \textbf{Mean}
& \textbf{0.50} & \textbf{Mean}
& \textbf{0.50} & \textbf{Mean}
& \textbf{0.50} & \textbf{Mean} \\

\midrule
\multicolumn{17}{c}{\textbf{Open-set Specialized Detectors}} \\
\midrule

Grounding DINO~\citep{groundingdino2024}
& 36.12 & 26.56
& \second{93.27} & \second{81.65}
& 46.78 & 40.93
& 42.36 & 30.19
& 10.53 & 7.00
& 39.86 & \second{27.70}
& 25.25 & 18.51
& 12.45 & 4.26 \\

\midrule
\multicolumn{17}{c}{\textbf{Closed-set Specialized Detectors}} \\
\midrule

DETR-R50~\citep{detr2020}
& 31.03 & 21.95
& 85.21 & 76.70
& 23.29 & 19.38
& 39.21 & 26.52
& 6.11 & 3.94
& 35.99 & 22.74
& 16.76 & 12.41
& 6.07 & 1.92 \\

EdgeCrafter~\citep{edgecrafter2026}
& 63.48 & 48.42
& 89.32 & 71.35
& 79.39 & 64.77
& 84.25 & 59.17
& 2.05 & 1.59
& 0.04 & 0.03
& 73.90 & 52.80
& 0.00 & 0.00 \\

YOLO26~\citep{yolo262026}
& 62.39 & 49.92
& 87.47 & 75.70
& 81.32 & 67.40
& 83.22 & 60.89
& 0.06 & 0.06
& 0.00 & 0.00
& 61.49 & 49.14
& 0.00 & 0.00 \\

LW-DETR-Object365~\citep{lwdetr2024}
& \second{72.35} & \best{55.31}
& 87.56 & 77.40
& \best{89.91} & \best{76.18}
& \best{87.26} & \best{62.80}
& 0.66 & 0.43
& 0.00 & 0.00
& \best{81.85} & \best{64.18}
& 0.00 & 0.00 \\

LW-DETR-DINOv2~\citep{lwdetr2024}
& 71.16 & \second{54.25}
& 85.88 & 75.42
& \second{89.13} & \second{74.66}
& \second{86.49} & \second{61.61}
& 0.60 & 0.47
& 0.00 & 0.00
& \second{78.88} & \second{58.62}
& 0.00 & 0.00 \\

\midrule
\multicolumn{17}{c}{\textbf{Vision-Language Models}} \\
\midrule

DeepSeek-VL2-Small~\citep{deepseekvl22024}
& 5.55 & 1.94
& 6.16 & 2.85
& 49.10 & 34.23
& 12.04 & 5.84
& 7.48 & 3.66
& 22.32 & 5.27
& 25.65 & 18.70
& 1.64 & 0.68 \\

MiMo-VL-7B~\citep{mimovl2025}
& 7.06 & 1.89
& 8.58 & 2.83
& 38.75 & 20.70
& 18.78 & 8.21
& 2.47 & 0.56
& 29.12 & 7.51
& 24.34 & 12.35
& 1.82 & 0.47 \\

LocateAnything-3B~\citep{locateanything2026}
& 7.47 & 4.55
& 6.16 & 3.01
& 41.69 & 32.90
& 21.60 & 10.95
& 9.34 & 6.60
& 21.78 & 6.36
& 33.37 & 25.29
& 7.11 & 4.63 \\

Ovis2.5-2B~\citep{ovis252025}
& 18.72 & 7.09
& 21.73 & 6.39
& 43.47 & 20.22
& 11.17 & 3.25
& 7.02 & 1.58
& 23.34 & 5.39
& 48.45 & 22.55
& 3.14 & 6.12 \\

Rex-Omni-3B~\citep{rexomni2026}
& 12.68 & 7.57
& 5.43 & 2.80
& 35.32 & 28.77
& 33.58 & 22.55
& 8.23 & 4.75
& 23.60 & 6.12
& 37.39 & 26.82
& 7.74 & 4.63 \\

GPT-5.5
& 7.65 & 3.17
& 41.44 & 16.57
& 65.62 & 41.88
& 38.36 & 20.65
& 15.53 & 6.27
& \second{56.63} & 23.81
& 36.96 & 17.83
& 12.11 & 4.12 \\

Qwen3-VL-4B~\citep{qwen3vl2025}
& 32.47 & 20.84
& 21.54 & 12.09
& 79.93 & 61.41
& 39.69 & 25.00
& 20.87 & 13.57
& 33.41 & 9.63
& 67.25 & 42.26
& 17.10 & 13.07 \\

Qwen3-VL-32B~\citep{qwen3vl2025}
& 26.72 & 23.41
& 24.00 & 18.53
& 79.90 & 65.20
& 42.97 & 30.17
& \best{37.40} & \best{25.05}
& 20.66 & 6.31
& 58.11 & 36.71
& \second{23.17} & \best{19.00} \\

Qwen3-VL-8B~\citep{qwen3vl2025}
& 32.69 & 24.08
& 33.31 & 24.29
& 81.44 & 66.68
& 43.53 & 29.59
& \second{24.54} & \second{18.99}
& 30.82 & 9.86
& 63.84 & 41.45
& 19.28 & \second{16.50} \\

\midrule

\textbf{LocAnyMed}
& \best{73.92} & 53.47
& \best{97.09} & \best{92.10}
& 85.91 & 70.22
& 83.37 & 54.96
& 11.35 & 6.02
& \best{57.92} & \best{35.24}
& 78.44 & 52.52
& \best{23.51} & 11.02 \\

\bottomrule
\end{tabular}%
}
\end{table*}

\subsection{Adaptation is consistent across modalities}

Table~\ref{supp:tab:modality_results} reports modality-wise performance on the \same{} and \cross{} splits. On the \same{} split, adapting LocateAnything-3B to medical data produces substantial improvements for all four modalities. Compared with LocateAnything-3B, LocAnyMed increases F1 Mean from 4.55 to 53.47 on CT, from 3.01 to 92.10 on ultrasound, from 32.90 to 70.22 on optical images, and from 10.95 to 54.96 on X-ray. The corresponding absolute gains are 48.92, 89.09, 37.32, and 44.01 percentage points, respectively. Thus, the overall improvement is not driven by a single high-performing modality.

The relative difficulty of the modalities also changes after adaptation. LocateAnything-3B performs best on optical images but remains weak on CT and ultrasound. After medical adaptation, ultrasound becomes the strongest same-set modality, reaching 97.09 F1@0.50 and 92.10 F1 Mean. LocAnyMed also achieves the highest CT F1@0.50 among the compared methods and remains competitive on optical and X-ray images. However, specialized closed-set detectors retain higher strict-IoU performance on some modalities, particularly optical and X-ray, showing that medical adaptation does not eliminate all differences between generative grounding and dedicated detection models.

Cross-set evaluation reveals a substantially different pattern. LocAnyMed retains 74.8\% of its same-set F1 Mean on optical images, compared with 38.3\% on ultrasound, 20.1\% on X-ray, and only 11.3\% on CT. Relative to LocateAnything-3B, cross-set F1 Mean improves by 28.88 percentage points on ultrasound, 27.23 points on optical images, and 6.39 points on X-ray. On CT, F1@0.50 increases from 9.34 to 11.35, whereas F1 Mean decreases slightly from 6.60 to 6.02. This discrepancy suggests that adaptation improves coarse CT localization without producing a corresponding improvement at stricter IoU thresholds.

\begin{figure*}[t]
\centering
\includegraphics[
    width=\textwidth,
    trim={5.5cm 6cm 5.5cm 2.5cm},
    clip
]{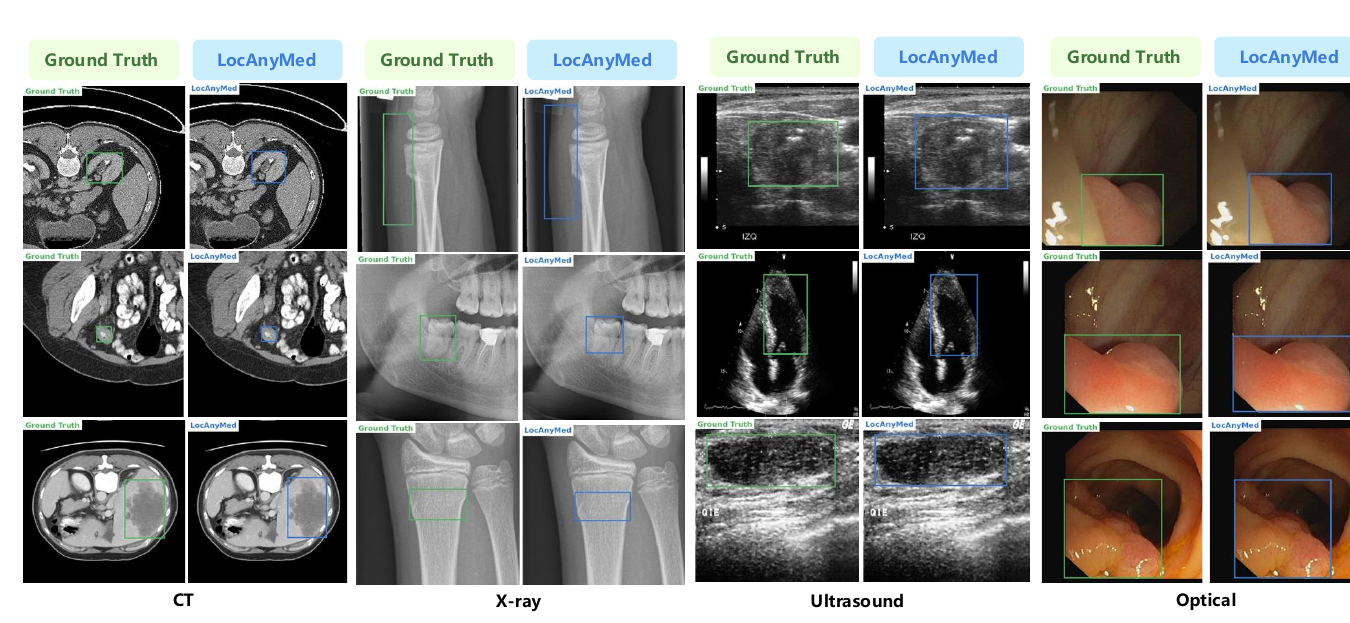}
\caption{Qualitative grounding results of LocAnyMed across four medical imaging modalities. Ground-truth boxes are shown in green, and model predictions are shown in blue.}
\label{supp:fig:qualitative-grounding}
\end{figure*}

The baseline comparison further illustrates the modality dependence of cross-source transfer. Closed-set detectors such as LW-DETR, YOLO26, and EdgeCrafter retain strong cross-set performance on optical images but collapse to nearly zero on ultrasound and X-ray. In contrast, LocAnyMed maintains non-zero performance across all four modalities, achieves the best cross-set ultrasound results, and obtains the highest X-ray F1@0.50. Qwen3-VL-32B remains stronger on CT F1 Mean and X-ray F1 Mean, while the LW-DETR variants retain an advantage on optical F1 Mean. These results show that no single method dominates every modality and that aggregate scores alone can obscure substantial differences in cross-source generalization. Figure~\ref{supp:fig:qualitative-grounding} presents representative
box-grounding predictions across CT, X-ray, ultrasound, and optical images.

\subsection{Checkpoint selection and restricted adaptation}

\begin{table}[t]
\centering
\caption{Dense full-parameter checkpoint curve on the complete sets (\%).}
\label{supp:tab:dense-curve}
\footnotesize
\setlength{\tabcolsep}{3.6pt}
\renewcommand{\arraystretch}{0.95}
\begin{tabular}{@{}rrrrr@{}}
\toprule
Step & Same F1@IoU 0.50 & Same Mean & Cross F1@IoU 0.50 & Cross Mean \\
\midrule
500  & 84.98 & 66.34 & 33.22 & \best{20.38} \\
1000 & 85.02 & 68.41 & 31.83 & 18.81 \\
1500 & \best{85.59} & \best{69.15} & \best{33.63} & 19.12 \\
2000 & 85.13 & 68.66 & 31.72 & 18.16 \\
2500 & 84.23 & 67.97 & 31.67 & 18.17 \\
\bottomrule
\end{tabular}
\end{table}

The dense curve in Table~\ref{supp:tab:dense-curve} supports selecting step 1500: \same{} F1 Mean peaks there, while later checkpoints lose both \same{} and \cross{} F1 Mean.  Under the stream-packed MTP loader, these steps are
optimization checkpoints rather than exact epochs.  The late decline should therefore be interpreted as checkpoint-level evidence that continued updates move the model toward the dominant training distribution, not as a claim about repeated full passes over every raw record.

Table~\ref{supp:tab:adapter-curves} adds the checkpoint curves behind the parameter-restricted rows in the main paper.  Both adapters improve rapidly, but visual-FFN-only adaptation saturates by 300--400 steps and projector-only adaptation saturates by 300 steps, remaining below full-parameter adaptation.

\begin{table}[t]
\centering
\caption{Restricted-adaptation F1 Mean curves on complete sets (\%).}
\label{supp:tab:adapter-curves}
\begin{tabular}{lrrr}
\toprule
Trainable parameters & Step & Same & Cross \\
\midrule
Projector only & 100 & 35.58 & 8.54 \\
& 300 & \best{42.18} & \best{12.62} \\
& 500 & 42.18 & 12.07 \\
\midrule
Visual FFN $fc1$ only & 100 & 46.67 & 6.46 \\
& 200 & 52.01 & 6.88 \\
& 300 & 56.88 & 10.47 \\
& 400 & \best{57.14} & \best{11.06} \\
& 500 & 56.45 & 10.96 \\
\bottomrule
\end{tabular}
\end{table}

\subsection{Unified output types and class imbalance}

Point supervision represents only 16,262 of the 209,910 training records (7.7\%), while absent-target queries account for 48,397 of the 193,648 detection records (25.0\%). Despite this imbalance, Table~\ref{supp:tab:auxiliary} shows that medical adaptation learns both point localization and explicit target rejection without separate task-specific losses. Dense SFT raises PCK@0.05 from 1.71 to 77.00 and negative-query accuracy from 9.67 to 92.92. The MoE variants remain within 1.22 points of one another on PCK@0.05 and within 0.70 points on negative-query accuracy, indicating that these capabilities are not sensitive to a particular expert allocation.

Table~\ref{supp:tab:point-main-curves} shows that point localization is learned early. Dense SFT reaches 77.36 PCK@0.05 after 500 steps and peaks at step 1000 with 78.16 PCK@0.05, 82.50 PCK@0.10, and an NND of 2.24. Later checkpoints remain close to this optimum, suggesting that point performance saturates before the end of training. The MoE models closely match dense SFT, whereas projector-only and visual-FFN-only tuning remain substantially weaker. At step 500, they reach 52.60 and 66.34 PCK@0.05, respectively, showing that full adaptation is important for accurate point prediction.

\begin{table}[t]
\centering
\caption{\same{} point and absent-target performance (\%).}
\label{supp:tab:auxiliary}
\footnotesize
\setlength{\tabcolsep}{5pt}
\renewcommand{\arraystretch}{0.95}
\begin{tabular}{@{}lrrr@{}}
\toprule
Model & PCK@0.05 & PCK@.10 & Neg. acc. \\
\midrule
LocateAnything-3B
& 1.71 & 8.02 & 9.67 \\

Dense SFT-1500
& 77.00 & 81.49 & 92.92 \\

MoE 3:1-1500
& 77.29 & \best{81.85} & 93.13 \\

MoE 2:2-1500
& 76.17 & 80.52 & \best{93.29} \\

MoE 7:1-1500
& \best{77.39} & 81.66 & 92.59 \\

MoE 2:2 + rationale-100
& 75.13 & 79.75 & 93.22 \\
\bottomrule
\end{tabular}
\end{table}

Table~\ref{supp:tab:point-rationale-curves} further separates learning point localization from preserving it. Training CoT directly from the official checkpoint reaches only 8.66 PCK@0.05 because CoT-20K contains no point records. In contrast, CoT continuation from medically adapted checkpoints retains approximately 75--76 PCK@0.05. NND nevertheless increases during continuation, indicating mild degradation that is not fully reflected by threshold-based PCK. Multi-scale continuation can recover strong point performance, but the grouped-stage results vary from 71.22 to 77.41 PCK@0.05, making checkpoint selection more important. These experiments therefore show that direct point supervision is necessary, while subsequent rationale training largely preserves, but does not improve, the learned point capability. External detector baselines are omitted because they do not generate point-coordinate sequences.

\begin{table}[t]
\centering
\caption{Complete \same{} point-localization results for medical SFT, MoE, and parameter-restricted variants. All rows are evaluated on 2,800 point records with 6,144 point instances. Values are percentages; lower NND is better.
}
\label{supp:tab:point-main-curves}
\footnotesize
\setlength{\tabcolsep}{2.5pt}
\renewcommand{\arraystretch}{0.96}

\begin{tabular}{@{}p{0.52\columnwidth}@{\hspace{8pt}}r@{\hspace{8pt}}r@{\hspace{8pt}}r@{}}
\toprule
Model/checkpoint & PCK@0.05 & PCK@.10 & NND \\
\midrule

\multicolumn{4}{l}{\textbf{Base}} \\
LocateAnything-3B
& 1.71 & 8.02 & 55.53 \\

\midrule
\multicolumn{4}{l}{\textbf{Dense SFT}} \\
Step 500
& 77.36 & 81.25 & 2.28 \\
Step 1000
& \best{78.16} & \best{82.50} & \best{2.24} \\
Step 1500
& 77.00 & 81.49 & 2.41 \\
Step 2000
& 76.81 & 81.18 & 2.40 \\
Step 2500
& 77.41 & 82.06 & 2.37 \\

\midrule
\multicolumn{4}{l}{\textbf{MoE SFT}} \\
3:1, step 1500
& 77.29 & \best{81.85} & \best{2.29} \\
2:2, step 1500
& 76.17 & 80.52 & 2.39 \\
7:1, step 1500
& \best{77.39} & 81.66 & 2.39 \\

\midrule
\multicolumn{4}{l}{\textbf{Projector only}} \\
Step 100
& 27.39 & 41.89 & 40.12 \\
Step 300
& 48.84 & 66.05 & 7.93 \\
Step 500
& \best{52.60} & \best{68.80} & \best{7.32} \\

\midrule
\multicolumn{4}{l}{\textbf{Visual FFN \(fc1\) only}} \\
Step 100
& 21.09 & 24.35 & 65.56 \\
Step 200
& 44.55 & 50.21 & 33.86 \\
Step 300
& 66.31 & \best{73.16} & 12.17 \\
Step 400
& 66.26 & 73.06 & 11.70 \\
Step 500
& \best{66.34} & 73.01 & \best{10.15} \\

\bottomrule
\end{tabular}
\end{table}

\begin{table}[t]
\centering
\caption{Complete \same{} point-localization results for rationale and multi-scale continuation. CoT-20K contains no point records. Values are percentages; lower NND is better.}
\label{supp:tab:point-rationale-curves}
\footnotesize
\setlength{\tabcolsep}{2.5pt}
\renewcommand{\arraystretch}{0.96}

\begin{tabular}{@{}p{0.52\columnwidth}@{\hspace{8pt}}r@{\hspace{8pt}}r@{\hspace{8pt}}r@{}}
\toprule
Model/checkpoint & PCK@0.05 & PCK@.10 & NND \\
\midrule

\multicolumn{4}{l}{\textbf{CoT from official}} \\
Step 500
& \best{8.66} & \best{15.93} & \best{62.73} \\
Step 1000
& 6.93 & 12.84 & 69.95 \\
Step 1500
& 7.06 & 12.45 & 70.74 \\
Step 2000
& 7.28 & 12.60 & 70.33 \\
Step 2500
& 7.41 & 12.86 & 70.03 \\

\midrule
\multicolumn{4}{l}{\textbf{Dense-1500 + CoT}} \\
Step 100
& 76.24 & 80.79 & \best{2.99} \\
Step 200
& 76.27 & 80.75 & 3.36 \\
Step 300
& 76.29 & 80.97 & 3.41 \\
Step 400
& \best{76.42} & \best{81.17} & 3.38 \\
Step 500
& 75.98 & 80.62 & 3.52 \\

\midrule
\multicolumn{4}{l}{\textbf{MoE 2:2-1500 + CoT}} \\
Step 100
& 75.13 & 79.75 & \best{2.77} \\
Step 200
& 75.75 & 80.00 & 2.92 \\
Step 300
& 76.03 & \best{80.42} & 2.89 \\
Step 400
& \best{76.11} & \best{80.42} & 3.18 \\
Step 500
& 75.81 & 80.05 & 3.38 \\

\midrule
\multicolumn{4}{l}{\textbf{MoE 2:2-1500 + last-3 MS + CoT}} \\
Step 100
& 74.97 & 79.31 & 3.00 \\
Step 200
& 75.54 & 80.03 & \best{2.82} \\
Step 300
& 76.24 & 80.47 & 2.92 \\
Step 400
& 76.07 & 80.27 & 2.93 \\
Step 500
& \best{76.32} & \best{80.52} & 2.95 \\

\midrule
\multicolumn{4}{l}{\textbf{Grouped-stage MS}} \\
Step 100
& 76.50 & \best{82.01} & 2.82 \\
Step 200
& 71.22 & 75.88 & 3.00 \\
Step 300
& \best{77.41} & 81.48 & 2.39 \\
Step 400
& 74.80 & 78.94 & 2.47 \\
Step 500
& 76.94 & 81.22 & \best{2.33} \\

\bottomrule
\end{tabular}
\end{table}

Micro- and macro-averaging answer different questions.  For Dense SFT-1500, \same{} F1@IoU 0.50/Mean are 85.59/69.15 micro versus 76.35/57.10 macro over 63 detection categories.  On \cross{} they are 33.63/19.12 micro versus 18.39/10.82 macro over 20 categories.  The lower macro scores expose long-tail categories that are hidden by record-weighted aggregation; we retain micro-F1 as the main metric because the benchmark unit is an image-query record, and report macro-F1 here as a complementary imbalance diagnostic.

\subsection{The preferred ratio depends on modality and source}

Table~\ref{supp:tab:moe-modality} disaggregates the three MoE allocations at the same 1500-step checkpoint. No allocation performs best across every modality and evaluation split. On the \same{} split, the 3:1 allocation gives the highest CT F1 Mean, the 2:2 allocation performs best on optical images, and the 7:1 allocation gives the strongest ultrasound and X-ray results. However, the same-set differences are relatively small: the gap between the best and worst allocation is only 0.72 points on CT, 1.69 points on optical images, 0.26 points on ultrasound, and 0.86 points on X-ray. Thus, all three allocations provide broadly similar in-distribution performance.

The allocation choice has a larger effect under cross-source shift. The 3:1 allocation performs best on CT, exceeding 7:1 by 2.84 points, whereas 2:2 performs best on X-ray, exceeding 3:1 by 3.06 points. The 7:1 allocation is strongest on both optical and ultrasound data, although 2:2 remains close, trailing by 1.10 and 0.06 points, respectively. These results indicate that cross-source performance is more sensitive to how visual capacity is divided between shared and modality-specific branches.

The pattern also suggests different capacity requirements across modalities. The larger shared branch in the 7:1 allocation appears beneficial for ultrasound and optical transfer, while the balanced 2:2 allocation, which assigns more capacity to the modality-specific branch, is more effective for the shifted X-ray distribution. CT instead favors the intermediate 3:1 allocation. Consequently, the effect of expert allocation is better interpreted as a modality-dependent capacity trade-off than as evidence for one universally optimal ratio. Because several differences are small, these results should be treated as comparative trends rather than definitive rankings without repeated-run variance estimates.

\begin{table}[t]
\centering
\caption{Per-modality F1 Mean for MoE allocation at step 1500 (\%).}
\label{supp:tab:moe-modality}
\small
\setlength{\tabcolsep}{5pt}
\renewcommand{\arraystretch}{1.05}

\begin{tabular}{@{}lrrr@{}}
\toprule
Modality & 3:1 & 2:2 & 7:1 \\
\midrule

\multicolumn{4}{c}{\textbf{\same{}}} \\
\cmidrule(lr){1-4}

CT
& \best{53.76} & 53.42 & 53.04 \\
Optical
& 69.92 & \best{70.65} & 68.96 \\
Ultrasound
& 92.00 & 91.82 & \best{92.08} \\
X-ray
& 54.93 & 54.36 & \best{55.22} \\

\midrule

\multicolumn{4}{c}{\textbf{\cross{}}} \\
\cmidrule(lr){1-4}

CT
& \best{6.59} & 5.84 & 3.75 \\
Optical
& 49.73 & 50.63 & \best{51.73} \\
Ultrasound
& 34.92 & 35.50 & \best{35.56} \\
X-ray
& 9.10 & \best{12.16} & 11.74 \\
\bottomrule
\end{tabular}
\end{table}

\section{Rationale Supervision}

\subsection{Distillation protocol and quality control}
The exact CoT subset contains 20,000 unique image hashes: 6,121 CT, 6,120 ultrasound, 6,120 X-ray, and 1,639 optical records, with 15,001 positive and 4,999 negative queries. Point records are excluded. The teacher used for the completed run is \texttt{gpt-5.5-xhigh}. For positive records it receives the complete image and a target-region contact sheet, plus modality, canonical target, presence status, and instance count. For negatives it receives only the complete image and metadata. It never receives the original user query or numeric coordinates.

\begin{table}[t]
\centering
\caption{Teacher prompt template for CoT-20K distillation.  Braced fields are filled from the manifest; one modality clause is appended according to the record metadata.}
\label{supp:tab:cot-prompt}
\scriptsize
\begin{tabularx}{\columnwidth}{p{0.20\columnwidth}Y}
\toprule
Field & Prompt content \\
\midrule
System & Generate a concise visual-evidence rationale for medical image localization SFT.  Use only the supplied image(s) and metadata.  Write one
English rationale of 20--45 words.  Do not predict, repeat, estimate, transform, or discuss coordinates.  Do not mention ground truth, annotations, bounding boxes, crops, panels, datasets, prompts, or hidden reasoning. Do not invent invisible findings.  Return valid JSON only:
\texttt{\{"rationale": "..."\}}. \\
\midrule
Positive & \texttt{Task: detection}.  \texttt{Target: \{target\}}.
\texttt{Target status: present}.  \texttt{Annotated instance count:
\{instance\_count\}}.  The first image is the complete medical image; the second contains target-region details only to clarify appearance. Describe the target-specific appearance, approximate relative location, and relationship to nearby visible structures. \\
\midrule
Negative & \texttt{Task: detection}.  \texttt{Target: \{target\}}.
\texttt{Target status: absent}.  Only the complete image is provided.  Begin
with: \texttt{No convincing visual evidence of the requested target is
visible.}  Then describe which expected target-specific appearance is missing,
without inventing an alternative finding. \\
\midrule
Modality & CT uses attenuation, contour, boundary, and nearby anatomy; X-ray uses density, lucency, cortical continuity, alignment, and projection-space relationships; ultrasound uses echogenicity, echotexture, shape, margin, orientation, internal pattern, and posterior acoustic appearance; optical imaging uses color, surface texture, contour, protrusion, depression, vascular pattern, and surrounding mucosal or surface context. \\
\bottomrule
\end{tabularx}
\end{table}

Table~\ref{supp:tab:cot-prompt} gives the prompt template used for teacher distillation. The common prompt requires one 20--45 word English rationale grounded in visible appearance, approximate relative location, and neighboring structures. Modality-specific clauses constrain the vocabulary to CT attenuation, radiographic density, sonographic appearance, or optical color/surface cues. Negative rationales must begin with a fixed absence statement and describe the expected evidence that is missing.  The teacher returns JSON containing only the rationale. The exporter then appends the original assistant target byte-for-byte, so the teacher cannot alter \texttt{<ref>} or coordinates.

\begin{table}[t]
\centering
\caption{Rationale-supervision ablation on the complete sets.  Values are F1 Mean (\%).}
\label{supp:tab:cot-ablation}
\begin{tabular}{lcc}
\toprule
Variant & \shortstack{Same-set\\Det. Val.} & \shortstack{Cross-set\\Det. Test} \\
\midrule
Official 3B + CoT & 56.76 & 14.67 \\
Dense-1500 + CoT & \best{69.05} & 20.10 \\
MoE 2:2-500 + CoT & 66.58 & 20.28 \\
MoE 2:2-1500 + CoT & 68.96 & \best{21.34} \\
\bottomrule
\end{tabular}
\end{table}

Automated validation enforces valid JSON, the word-count interval, forbidden coordinate/annotation/layout terms, positive/negative consistency, a clear negation cue for negatives, unique image hashes, and exact preservation of the original grounding suffix. All 20,000 records passed; 19,936 succeeded on the first API attempt, 57 on the second, and seven on the third. The 100-record pilot was manually reviewed before the full run, followed by manual spot inspection of the final corpus.

\subsection{CoT refines}
Table~\ref{supp:tab:cot-ablation} controls the checkpoint from which CoT SFT starts. Training the official general-domain checkpoint directly on only 20K rationale records is substantially weaker than first learning medical perception from the 200K corpus.  Starting from the undertrained MoE-500 checkpoint also remains below the MoE-1500 continuation.  CoT therefore acts as a refinement stage, not a substitute for broad medical grounding supervision.  The table reports the selected checkpoint for each continuation experiment and does not treat continuation step count as an independent comparison axis.

For the selected MoE 2:2 SFT-1500 initialization, later continuation checkpoints do not improve the result: \same{} F1 Mean changes from 68.96 to
68.88, 68.25, 68.07, and 67.97, while \cross{} F1 Mean changes from 21.34 to 21.02, 19.96, 19.86, and 19.93.  Because continuation uses the same
stream-packed MTP loader, these are checkpoint-selection observations rather than exact epoch-level claims.

\section{Multi-scale Fusion}

We evaluate two placements of multi-scale fusion under different training regimes. Grouped-stage fusion is trained from the official dense checkpoint on the full 200K corpus and combines the outputs of MoonViT blocks 9, 18, and 27. Last-three-block fusion instead starts from MoE 2:2-1500 and is evaluated during CoT continuation. Because the two settings use different initial checkpoints and training data, their absolute scores should not be interpreted as a direct architectural comparison; the relevant comparison is within each training regime.

As shown in Table~\ref{supp:tab:multiscale-ablation}, grouped-stage fusion improves \same{} F1 Mean steadily from 59.64 at step 100 to 66.89 at step 500, an increase of 7.25 points. Its \cross{} performance does not follow the same trend: it peaks at 19.42 after 200 steps, decreases to 18.32 at step 400, and ends at 19.06. Thus, aggregating widely separated visual stages consistently benefits in-distribution fitting but does not guarantee stronger cross-source generalization.

\begin{table}[t]
\centering
\caption{Multi-scale fusion ablation under two training regimes. Values are F1 Mean (\%).}
\label{supp:tab:multiscale-ablation}
\small
\setlength{\tabcolsep}{5pt}
\renewcommand{\arraystretch}{1.05}

\begin{tabular}{@{}lrrr@{}}
\toprule
Configuration & Step & \same{} & \cross{} \\
\midrule

\multicolumn{4}{l}{\textit{Full 200K SFT from the official checkpoint}} \\

\multirow{5}{*}{%
  \shortstack[l]{Grouped stages\\(9/18/27)}
}
& 100 & 59.64 & 17.10 \\
& 200 & 62.33 & \best{19.42} \\
& 300 & 64.39 & 19.03 \\
& 400 & 66.22 & 18.32 \\
& 500 & \best{66.89} & 19.06 \\

\midrule

\multicolumn{4}{l}{%
  \textit{100-step CoT continuation from MoE 2:2-1500}
} \\

No fusion
& 100 & 68.96 & 21.34 \\

\shortstack[l]{Last three\\(25/26/27)}
& 100 & \best{69.04} & \best{21.51} \\

\bottomrule
\end{tabular}
\end{table}

The last-three-block experiment provides a controlled comparison under the same 100-step CoT continuation. Adding blocks 25, 26, and 27 increases
\same{} F1 Mean from 68.96 to 69.04 and \cross{} F1 Mean from 21.34 to 21.51. These gains are positive but small, indicating that late-layer fusion
acts as a modest refinement rather than a dominant source of improvement. Longer continuation does not strengthen this advantage: \same{} F1 Mean
decreases to 68.73, 68.21, 68.16, and 68.12 at steps 200--500, while \cross{} F1 Mean changes to 20.79, 20.13, 20.14, and 20.25. Overall, multi-scale fusion provides a small early benefit, but its effectiveness depends on placement, training stage, and checkpoint selection.

\section{Inference Efficiency}

Table~\ref{supp:tab:efficiency} reports inference efficiency on the complete 45,437-record \same{} set. All measurements use batch size one on NVIDIA H20
GPUs. End-to-end FPS counts completed queries per second, whereas token FPS counts generated tokens per second. Both are computed from recorded per-sample generation time and should not be interpreted as aggregate four-GPU wall-clock throughput. The latency columns report per-query percentiles, and peak memory is measured independently for each GPU replica.

Medical SFT improves completed-query throughput from 2.09 to 3.20 FPS while reducing median latency from 0.392 to 0.334 seconds. This improvement is partly explained by the shorter output length: Dense SFT generates 10.87 tokens per query, compared with 16.92 for LocateAnything-3B. The similar token rates of the two models indicate that the gain primarily comes from producing shorter and more task-specific responses rather than from substantially faster token generation.

The three MoE allocations remain close to Dense SFT in both speed and latency. Their end-to-end throughput ranges from 3.10 to 3.16 FPS, compared with 3.20 FPS for the dense model, while median latency remains between 0.333 and 0.338 seconds. Relative to Dense SFT, the additional peak memory is approximately 101--383 MiB, with the balanced 2:2 allocation using the most memory. This modest overhead is consistent with storing multiple modality-specific branches while activating only one branch for each input.

\begin{table*}[t]
\centering
\caption{
Inference efficiency on the complete \same{} set using NVIDIA H20 GPUs at batch size one. Latency is measured per query, and memory is reported per GPU replica.}
\label{supp:tab:efficiency}
\small
\setlength{\tabcolsep}{6pt}
\renewcommand{\arraystretch}{1.08}

\begin{tabular}{@{}lrrrrrr@{}}
\toprule
\multirow{2}{*}{\textbf{Model}}
& \multicolumn{4}{c}{\textbf{Speed and latency}}
& \multicolumn{2}{c}{\textbf{Memory and output}} \\
\cmidrule(lr){2-5}
\cmidrule(l){6-7}
& \textbf{E2E FPS}
& \textbf{Token FPS}
& \textbf{p50 (s)}
& \textbf{p95 (s)}
& \textbf{Peak MiB}
& \textbf{Tokens/query} \\
\midrule

\multicolumn{7}{l}{\textit{Short-output grounding models}} \\

LocateAnything-3B
& 2.09 & 35.31 & 0.392 & 0.622 & 7,834 & 16.92 \\

Dense SFT-1500
& 3.20 & 34.76 & 0.334 & 0.562 & 7,829 & 10.87 \\

MoE 3:1-1500
& 3.16 & 34.39 & 0.333 & 0.565 & 8,032 & 10.90 \\

MoE 2:2-1500
& 3.14 & 34.12 & 0.338 & 0.556 & 8,212 & 10.87 \\

MoE 7:1-1500
& 3.10 & 33.86 & 0.338 & 0.560 & 7,930 & 10.93 \\

\midrule
\multicolumn{7}{l}{\textit{Rationale-generating models}} \\

MoE 2:2 + rationale-100
& 0.66 & 38.61 & 1.514 & 1.843 & 8,540 & 58.58 \\

MoE 2:2 + MS + rationale-100
& 0.66 & 39.45 & 1.502 & 1.823 & 8,540 & 59.69 \\

\bottomrule
\end{tabular}
\end{table*}

Rationale generation exhibits a different throughput profile. It increases the average output length from approximately 11 to 59 tokens and reduces completed-query throughput from 3.14 to 0.66 FPS. However, token throughput increases from 34.12 to 38.61 tokens per second. The resulting slowdown therefore arises mainly from generating longer answers rather than from a slower visual encoder or decoder. Adding multi-scale fusion to the rationale model leaves peak memory unchanged and produces nearly identical latency, indicating negligible additional deployment cost in this configuration.

The evaluator launches one independent process per GPU, assigns a disjoint JSONL shard to each process, and merges the parsed predictions after all
processes finish. Reported latency and FPS values exclude model loading, dataset scanning, file-system latency, and scheduler overhead. Consequently, they are intended for controlled model comparison rather than as hardware-independent deployment benchmarks. We also omit modality-wise speed as a primary result because differences across modalities mainly reflect image resolution, token count, and file-system variation. MoE routing itself is deterministic and does not introduce a learned router or additional generated tokens.

\clearpage
\bibliographystyle{plainnat}
\bibliography{LocAnyMed}

@inproceedings{nndetection2021,
  title={nnDetection: a self-configuring method for medical object detection},
  author={Baumgartner, Michael and J{\"a}ger, Paul F and Isensee, Fabian and Maier-Hein, Klaus H},
  booktitle={International conference on medical image computing and computer-assisted intervention},
  pages={530--539},
  year={2021},
  organization={Springer}
}

@inproceedings{medrpg2023,
  title={Medical phrase grounding with region-phrase context contrastive alignment},
  author={Chen, Zhihao and Zhou, Yang and Tran, Anh and Zhao, Junting and Wan, Liang and Ooi, Gideon Su Kai and Cheng, Lionel Tim-Ee and Thng, Choon Hua and Xu, Xinxing and Liu, Yong and others},
  booktitle={International Conference on Medical Image Computing and Computer-Assisted Intervention},
  pages={371--381},
  year={2023},
  organization={Springer}
}

@article{umedground2025,
  title={Uncertainty-aware medical diagnostic phrase identification and grounding},
  author={Zou, Ke and Bai, Yang and Liu, Bo and Chen, Yidi and Chen, Zhihao and Zhou, Yang and Yuan, Xuedong and Wang, Meng and Shen, Xiaojing and Cao, Xiaochun and others},
  journal={IEEE Transactions on Pattern Analysis and Machine Intelligence},
  year={2025},
  publisher={IEEE}
}

@misc{locateanything2026,
  title={LocateAnything: Fast and high-quality vision-language grounding with parallel box decoding},
  author={Wang, Shihao and Liu, Shilong and Kuang, Yuanguo and Wei, Xinyu and Liu, Yangzhou and Li, Zhiqi and Man, Yunze and Chen, Guo and Tao, Andrew and Liu, Guilin and others},
  journal={arXiv preprint arXiv:2605.27365},
  year={2026}
}

@article{medsam2024,
  title={Segment anything in medical images},
  author={Ma, Jun and He, Yuting and Li, Feifei and Han, Lin and You, Chenyu and Wang, Bo},
  journal={Nature communications},
  volume={15},
  number={1},
  pages={654},
  year={2024},
  publisher={Nature Publishing Group UK London}
}

@article{biomedparse2025,
  title={A foundation model for joint segmentation, detection and recognition of biomedical objects across nine modalities},
  author={Zhao, Theodore and Gu, Yu and Yang, Jianwei and Usuyama, Naoto and Lee, Ho Hin and Kiblawi, Sid and Naumann, Tristan and Gao, Jianfeng and Crabtree, Angela and Abel, Jacob and others},
  journal={Nature methods},
  volume={22},
  number={1},
  pages={166--176},
  year={2025},
  publisher={Nature Publishing Group US New York}
}

@inproceedings{m4oe2024,
  title={M4oe: A foundation model for medical multimodal image segmentation with mixture of experts},
  author={Jiang, Yufeng and Shen, Yiqing},
  booktitle={international conference on medical image computing and computer-assisted intervention},
  pages={621--631},
  year={2024},
  organization={Springer}
}

@misc{medmoe2025,
  title={Medmoe: modality-specialized mixture of experts for medical vision-language understanding},
  author={Chopra, Shivang and Sanchez-Rodriguez, Gabriela and Mao, Lingchao and Feola, Andrew J and Li, Jing and Kira, Zsolt},
  journal={arXiv preprint arXiv:2506.08356},
  year={2025}
}

@inproceedings{rexthinker2026,
  title={Rex-thinker: Grounded object referring via chain-of-thought reasoning},
  author={Jiang, Qing and Chen, Xingyu and Zeng, Zhaoyang and Yu, Junzhi and Zhang, Lei},
  journal={arXiv preprint arXiv:2506.04034},
  year={2025}
}

@misc{medclm2025,
  title={MedCLM: Learning to Localize and Reason via a CoT-Curriculum in Medical Vision-Language Models},
  author={Kim, Soo Yong and Cho, Suin and Yun, Vincent-Daniel and Hwang, Gyeongyeon},
  journal={arXiv preprint arXiv:2510.04477},
  year={2025}
}

@article{deeplesion2018,
  title={DeepLesion: automated mining of large-scale lesion annotations and universal lesion detection with deep learning},
  author={Yan, Ke and Wang, Xiaosong and Lu, Le and Summers, Ronald M},
  journal={Journal of medical imaging},
  volume={5},
  number={3},
  pages={036501--036501},
  year={2018},
  publisher={Society of Photo-Optical Instrumentation Engineers}
}

@inproceedings{mdetr2021,
  title={Mdetr-modulated detection for end-to-end multi-modal understanding},
  author={Kamath, Aishwarya and Singh, Mannat and LeCun, Yann and Synnaeve, Gabriel and Misra, Ishan and Carion, Nicolas},
  booktitle={Proceedings of the IEEE/CVF international conference on computer vision},
  pages={1780--1790},
  year={2021}
}

@inproceedings{glip2022,
  title={Grounded language-image pre-training},
  author={Li, Liunian Harold and Zhang, Pengchuan and Zhang, Haotian and Yang, Jianwei and Li, Chunyuan and Zhong, Yiwu and Wang, Lijuan and Yuan, Lu and Zhang, Lei and Hwang, Jenq-Neng and others},
  booktitle={Proceedings of the IEEE/CVF conference on computer vision and pattern recognition},
  pages={10965--10975},
  year={2022}
}

@inproceedings{owlvit2022,
  title={Simple open-vocabulary object detection},
  author={Minderer, Matthias and Gritsenko, Alexey and Stone, Austin and Neumann, Maxim and Weissenborn, Dirk and Dosovitskiy, Alexey and Mahendran, Aravindh and Arnab, Anurag and Dehghani, Mostafa and Shen, Zhuoran and others},
  booktitle={European conference on computer vision},
  pages={728--755},
  year={2022},
  organization={Springer}
}

@inproceedings{groundingdino2024,
  title={Grounding dino: Marrying dino with grounded pre-training for open-set object detection},
  author={Liu, Shilong and Zeng, Zhaoyang and Ren, Tianhe and Li, Feng and Zhang, Hao and Yang, Jie and Jiang, Qing and Li, Chunyuan and Yang, Jianwei and Su, Hang and others},
  booktitle={European conference on computer vision},
  pages={38--55},
  year={2024},
  organization={Springer}
}

@misc{kosmos22023,
  title={Grounding multimodal large language models to the world},
  author={Peng, Zhiliang and Wang, Wenhui and Dong, Li and Hao, Yaru and Huang, Shaohan and Ma, Shuming and Ye, Qixiang and Wei, Furu},
  booktitle={International Conference on Learning Representations},
  volume={2024},
  pages={51575--51598},
  year={2024}
}

@inproceedings{rexomni2026,
  title={Detect anything via next point prediction},
  author={Jiang, Qing and Huo, Junan and Chen, Xingyu and Xiong, Yuda and Zeng, Zhaoyang and Chen, Yihao and Ren, Tianhe and Yu, Junzhi and Zhang, Lei},
  journal={arXiv preprint arXiv:2510.12798},
  year={2025}
}

@inproceedings{medrov2026,
  title={MedROV: Towards Real-Time Open-Vocabulary Detection Across Diverse Medical Imaging Modalities},
  author={Sheikh, Tooba Tehreem and Lahoud, Jean and Anwer, Rao Muhammad and Khan, Fahad Shahbaz and Khan, Salman and Cholakkal, Hisham},
  booktitle={2026 IEEE/CVF Winter Conference on Applications of Computer Vision (WACV)},
  pages={8628--8638},
  year={2026},
  organization={IEEE}
}

@misc{vividmed2024,
  title={Vividmed: Vision language model with versatile visual grounding for medicine},
  author={Luo, Lingxiao and Tang, Bingda and Chen, Xuanzhong and Han, Rong and Chen, Ting},
  journal={arXiv preprint arXiv:2410.12694},
  year={2024}
}

@inproceedings{mimo2025,
  title={MIMO: A medical vision language model with visual referring multimodal input and pixel grounding multimodal output},
  author={Chen, Yanyuan and Xu, Dexuan and Huang, Yu and Zhan, Songkun and Wang, Hanpin and Chen, Dongxue and Wang, Xueping and Qiu, Meikang and Li, Hang},
  booktitle={Proceedings of the Computer Vision and Pattern Recognition Conference},
  pages={24732--24741},
  year={2025}
}

@misc{medplib2024,
  title={Towards a multimodal large language model with pixel-level insight for biomedicine},
  author={Huang, Xiaoshuang and Shen, Lingdong and Liu, Jia and Shang, Fangxin and Li, Hongxiang and Huang, Haifeng and Yang, Yehui},
  booktitle={Proceedings of the AAAI Conference on Artificial Intelligence},
  volume={39},
  number={4},
  pages={3779--3787},
  year={2025}
}

@article{unibiomed2026,
  title={A universal foundation model for grounded biomedical image interpretation},
  author={Wu, Linshan and Nie, Yuxiang and He, Sunan and Zhuang, Jiaxin and Luo, Luyang and Li, Tao and Xie, Zhuoyao and Chen, Dexuan and Zhao, Yinghua and Mahboobani, Neeraj and others},
  journal={Nature Communications},
  year={2026},
  publisher={Nature Publishing Group}
}

@inproceedings{biovil2022,
  title={Making the most of text semantics to improve biomedical vision--language processing},
  author={Boecking, Benedikt and Usuyama, Naoto and Bannur, Shruthi and Castro, Daniel C and Schwaighofer, Anton and Hyland, Stephanie and Wetscherek, Maria and Naumann, Tristan and Nori, Aditya and Alvarez-Valle, Javier and others},
  booktitle={European conference on computer vision},
  pages={1--21},
  year={2022},
  organization={Springer}
}

@misc{medground2026,
  title={MedGround: Bridging the Evidence Gap in Medical Vision-Language Models with Verified Grounding Data},
  author={Zhang, Mengmeng and Wu, Xiaoping and Luo, Hao and Wang, Fan and Lv, Yisheng},
  journal={arXiv preprint arXiv:2601.06847},
  year={2026}
}

@inproceedings{unimed2024,
  title={Uni-med: a unified medical generalist foundation model for multi-task learning via connector-MoE},
  author={Zhu, Xun and Hu, Ying and Mo, Fanbin and Li, Miao and Wu, Ji},
  journal={Advances in Neural Information Processing Systems},
  volume={37},
  pages={81225--81256},
  year={2024}
}

@misc{univgr12025,
  title={Univg-r1: Reasoning guided universal visual grounding with reinforcement learning},
  author={Bai, Sule and Li, Mingxing and Liu, Yong and Tang, Jing and Zhang, Haoji and Sun, Lei and Chu, Xiangxiang and Tang, Yansong},
  journal={arXiv preprint arXiv:2505.14231},
  year={2025}
}

@misc{gemex2024,
  title={Gemex: A large-scale, groundable, and explainable medical vqa benchmark for chest x-ray diagnosis},
  author={Liu, Bo and Zou, Ke and Zhan, Li-Ming and Lu, Zexin and Dong, Xiaoyu and Chen, Yidi and Xie, Chengqiang and Cao, Jiannong and Wu, Xiao-Ming and Fu, Huazhu},
  booktitle={Proceedings of the IEEE/CVF International Conference on Computer Vision},
  pages={21310--21320},
  year={2025}
}

@misc{bettereyes2026,
  title={Better eyes, better thoughts: Why vision chain-of-thought fails in medicine},
  author={Wu, Yuan and Yang, Zongxian and Qian, Jiayu and Gao, Songpan and Chen, Guanxing and Li, Qiankun and Huang, Yu-An and Huang, Zhi-An},
  journal={arXiv preprint arXiv:2603.06665},
  year={2026}
}

@article{msd2022,
  title={The medical segmentation decathlon},
  author={Antonelli, Michela and Reinke, Annika and Bakas, Spyridon and Farahani, Keyvan and Kopp-Schneider, Annette and Landman, Bennett A and Litjens, Geert and Menze, Bjoern and Ronneberger, Olaf and Summers, Ronald M and others},
  journal={Nature communications},
  volume={13},
  number={1},
  year={2022},
  publisher={Nature Portfolio}
}

@article{totalsegmentator2023,
  title={TotalSegmentator: robust segmentation of 104 anatomic structures in CT images},
  author={Wasserthal, Jakob and Breit, Hanns-Christian and Meyer, Manfred T and Pradella, Maurice and Hinck, Daniel and Sauter, Alexander W and Heye, Tobias and Boll, Daniel T and Cyriac, Joshy and Yang, Shan and others},
  journal={Radiology: Artificial Intelligence},
  volume={5},
  number={5},
  pages={e230024},
  year={2023},
  publisher={Radiological Society of North America}
}

@article{luna162017,
  title={Validation, comparison, and combination of algorithms for automatic detection of pulmonary nodules in computed tomography images: the LUNA16 challenge},
  author={Setio, Arnaud Arindra Adiyoso and Traverso, Alberto and De Bel, Thomas and Berens, Moira SN and Van Den Bogaard, Cas and Cerello, Piergiorgio and Chen, Hao and Dou, Qi and Fantacci, Maria Evelina and Geurts, Bram and others},
  journal={Medical image analysis},
  volume={42},
  pages={1--13},
  year={2017},
  publisher={Elsevier}
}

@inproceedings{samus2024,
  title={Beyond adapting SAM: Towards end-to-end ultrasound image segmentation via auto prompting},
  author={Lin, Xian and Xiang, Yangyang and Yu, Li and Yan, Zengqiang},
  booktitle={International Conference on Medical Image Computing and Computer-Assisted Intervention},
  pages={24--34},
  year={2024},
  organization={Springer}
}

@article{busbra2024,
  title={BUS-BRA: A breast ultrasound dataset for assessing computer-aided diagnosis systems},
  author={G{\'o}mez-Flores, Wilfrido and Gregorio-Calas, Maria Julia and Coelho de Albuquerque Pereira, Wagner},
  journal={Medical physics},
  volume={51},
  number={4},
  pages={3110--3123},
  year={2024},
  publisher={Wiley Online Library}
}

@article{grazpedwri2022,
  title={A pediatric wrist trauma X-ray dataset (GRAZPEDWRI-DX) for machine learning},
  author={Nagy, Eszter and Janisch, Michael and Hr{\v{z}}i{\'c}, Franko and Sorantin, Erich and Tschauner, Sebastian},
  journal={Scientific data},
  volume={9},
  number={1},
  pages={222},
  year={2022},
  publisher={Nature Publishing Group UK London}
}

@article{fracatlas2023,
  title={Fracatlas: A dataset for fracture classification, localization and segmentation of musculoskeletal radiographs},
  author={Abedeen, Iftekharul and Rahman, Md Ashiqur and Prottyasha, Fatema Zohra and Ahmed, Tasnim and Chowdhury, Tareque Mohmud and Shatabda, Swakkhar},
  journal={Scientific data},
  volume={10},
  number={1},
  pages={521},
  year={2023},
  publisher={Nature Publishing Group UK London}
}

@misc{dentex2023,
  title={Dentex: An abnormal tooth detection with dental enumeration and diagnosis benchmark for panoramic x-rays},
  author={Hamamci, Ibrahim Ethem and Er, Sezgin and Simsar, Enis and Yuksel, Atif Emre and Gultekin, Sadullah and Ozdemir, Serife Damla and Yang, Kaiyuan and Li, Hongwei Bran and Pati, Sarthak and Stadlinger, Bernd and others},
  journal={arXiv preprint arXiv:2305.19112},
  year={2023}
}

@inproceedings{kvasirseg2020,
  title={Kvasir-seg: A segmented polyp dataset},
  author={Jha, Debesh and Smedsrud, Pia H and Riegler, Michael A and Halvorsen, P{\aa}l and De Lange, Thomas and Johansen, Dag and Johansen, H{\aa}vard D},
  booktitle={International conference on multimedia modeling},
  pages={451--462},
  year={2019},
  organization={Springer}
}

@article{cvcclinicdb2015,
  title={WM-DOVA maps for accurate polyp highlighting in colonoscopy: Validation vs. saliency maps from physicians},
  author={Bernal, Jorge and S{\'a}nchez, F Javier and Fern{\'a}ndez-Esparrach, Gloria and Gil, Debora and Rodr{\'\i}guez, Cristina and Vilari{\~n}o, Fernando},
  journal={Computerized medical imaging and graphics},
  volume={43},
  pages={99--111},
  year={2015},
  publisher={Elsevier}
}

@article{ddr2019,
  title={Diagnostic assessment of deep learning algorithms for diabetic retinopathy screening},
  author={Li, Tao and Gao, Yingqi and Wang, Kai and Guo, Song and Liu, Hanruo and Kang, Hong},
  journal={Information Sciences},
  volume={501},
  pages={511--522},
  year={2019},
  publisher={Elsevier}
}

@article{drive2004,
  title={Ridge-based vessel segmentation in color images of the retina},
  author={Staal, Joes and Abr{\`a}moff, Michael D and Niemeijer, Meindert and Viergever, Max A and Van Ginneken, Bram},
  journal={IEEE transactions on medical imaging},
  volume={23},
  number={4},
  pages={501--509},
  year={2004},
  publisher={IEEE}
}

@article{stare2000,
  title={Locating blood vessels in retinal images by piecewise threshold probing of a matched filter response},
  author={Hoover, AD and Kouznetsova, Valentina and Goldbaum, Michael},
  journal={IEEE Transactions on Medical imaging},
  volume={19},
  number={3},
  pages={203--210},
  year={2000},
  publisher={IEEE}
}

@inproceedings{detr2020,
  title={End-to-end object detection with transformers},
  author={Carion, Nicolas and Massa, Francisco and Synnaeve, Gabriel and Usunier, Nicolas and Kirillov, Alexander and Zagoruyko, Sergey},
  booktitle={European conference on computer vision},
  pages={213--229},
  year={2020},
  organization={Springer}
}

@misc{yolo262026,
  title={Ultralytics YOLO26: unified real-time end-to-end vision models},
  author={Jocher, Glenn and Qiu, Jing and Liu, Mengyu and Lyu, Shuai and Akyon, Fatih Cagatay and Kalfaoglu, Muhammet Esat},
  journal={arXiv preprint arXiv:2606.03748},
  year={2026}
}

@misc{lwdetr2024,
  title={Lw-detr: A transformer replacement to yolo for real-time detection},
  author={Chen, Qiang and Su, Xiangbo and Zhang, Xinyu and Wang, Jian and Chen, Jiahui and Shen, Yunpeng and Han, Chuchu and Chen, Ziliang and Xu, Weixiang and Li, Fanrong and others},
  journal={arXiv preprint arXiv:2406.03459},
  year={2024}
}

@misc{edgecrafter2026,
  title={EdgeCrafter: Compact ViTs for Edge Dense Prediction via Task-Specialized Distillation},
  author={Liu, Longfei and Hou, Yongjie and Li, Yang and Wang, Qirui and Sha, Youyang and Yu, Yongjun and Wang, Yinzhi and Ru, Peizhe and Yu, Xuanlong and Shen, Xi},
  journal={arXiv preprint arXiv:2603.18739},
  year={2026}
}

@misc{deepseekvl22024,
  title={Deepseek-vl2: Mixture-of-experts vision-language models for advanced multimodal understanding},
  author={Wu, Zhiyu and Chen, Xiaokang and Pan, Zizheng and Liu, Xingchao and Liu, Wen and Dai, Damai and Gao, Huazuo and Ma, Yiyang and Wu, Chengyue and Wang, Bingxuan and others},
  journal={arXiv preprint arXiv:2412.10302},
  year={2024}
}

@misc{mimovl2025,
  title={Xiaomi MiMo-VL-Miloco Technical Report},
  author={Li, Jiaze and Chen, Jingyang and Qu, Yuxun and Xu, Shijie and Lin, Zhenru and Zhu, Junyou and Xu, Boshen and Tan, Wenhui and Fu, Pei and Ju, Jianzhong and others},
  journal={arXiv preprint arXiv:2512.17436},
  year={2025}
}

@misc{ovis252025,
  title={Ovis2. 5 technical report},
  author={Lu, Shiyin and Li, Yang and Xia, Yu and Hu, Yuwei and Zhao, Shanshan and Ma, Yanqing and Wei, Zhichao and Li, Yinglun and Duan, Lunhao and Zhao, Jianshan and others},
  journal={arXiv preprint arXiv:2508.11737},
  year={2025}
}

@misc{qwen3vl2025,
  title={Qwen3-vl technical report},
  author={Bai, Shuai and Cai, Yuxuan and Chen, Ruizhe and Chen, Keqin and Chen, Xionghui and Cheng, Zesen and Deng, Lianghao and Ding, Wei and Gao, Chang and Ge, Chunjiang and others},
  journal={arXiv preprint arXiv:2511.21631},
  year={2025}
}

\end{document}